\documentclass[letterpaper]{article} 
\usepackage[preprint]{aaai2027}  
\usepackage[hyphens]{url}  
\usepackage{graphicx} 
\usepackage{natbib}  
\usepackage{caption} 
\usepackage{algorithm}
\usepackage{algorithmic}
\usepackage{multibib}
\newcites{supp}{Supplementary References}

\usepackage{newfloat}
\usepackage{listings}
\DeclareCaptionStyle{ruled}{labelfont=normalfont,labelsep=colon,strut=off} 
\floatstyle{ruled}
\newfloat{listing}{tb}{lst}{}
\floatname{listing}{Listing}

\usepackage{booktabs}
\usepackage{amsmath}
\usepackage{amssymb}
\usepackage{multirow}
\usepackage{makecell}

\nocopyright

\title{Rethinking EEG-Based Disease Diagnosis: Decoupling Instance Representation Learning from Subject-Level Supervision}
\author{
    Zhiyuan Ma\textsuperscript{\rm 1}\equalcontrib,
    Zeyuan Li\textsuperscript{\rm 1}\equalcontrib,
    Zhiyi Lu\textsuperscript{\rm 1}\equalcontrib,
    Jiacheng Hao\textsuperscript{\rm 1},
    Youlang Du\textsuperscript{\rm 1},
    Zhen Jiang\textsuperscript{\rm 1},
    Xinche Zhang\textsuperscript{\rm 1},
    Yuhao Sun\textsuperscript{\rm 1},
    Xinke Shen\textsuperscript{\rm 2}\corresponding,
    Sen Song\textsuperscript{\rm 1}\corresponding
}
\affiliations{
    \textsuperscript{\rm 1}School of Biomedical Engineering, Tsinghua Medicine, Tsinghua University\\
    \textsuperscript{\rm 2}Department of Biomedical Engineering, Southern University of Science and Technology\\

    \{ma-zy25, zeyuan-l23, lzd25\}@mails.tsinghua.edu.cn, shenxk@sustech.edu.cn, songsen@tsinghua.edu.cn
}

\begin{document}

\maketitle

\begin{abstract}

EEG-based disease diagnosis requires one prediction per subject, yet common pipelines segment recordings into short instances, inherit the subject label for every instance, and train instance-level classifiers. This assumes that all instances provide equally reliable diagnostic evidence. Multiple instance learning (MIL) avoids inherited labels by treating each subject as a bag. However, EEG datasets contain far fewer subjects than instances, which can limit the quality of the representations learned by end-to-end MIL.
We propose \textbf{BridgeMIL}, a two-stage framework that decouples instance representation learning from subject-level supervision. Stage 1 pretrains the encoder without inherited instance labels by aligning temporally nearby windows and independently sampled within-subject sub-bags. Variance and covariance regularization prevent collapse and reduce redundancy without negative pairs. Stage 2 transfers the encoder to an attention-based MIL aggregator, applies supervision only to subject predictions, and limits representation drift through feature retention.
Across three EEG disease datasets and five representative backbones, BridgeMIL attains the highest mean accuracy in 14 of 15 dataset–backbone settings and an overall mean accuracy of 76.57\%, 4.28 percentage points higher than the strongest baseline. Further analyses reveal substantial variation in inherited-label reliability across instances, greater performance sensitivity to subject scarcity than to instance scarcity, and a more structured representation space with distinct subject-wise clusters and improved separation between diagnostic classes. Together, these findings underscore the importance of aligning supervision with the subject-level prediction objective while learning from abundant EEG instances without assigning disease labels to individual instances.

\end{abstract}


\section*{Introduction}
Deep learning has shown promise in EEG-based diagnosis of neurological and psychiatric disorders \cite{parsa2023eeg}. Most existing methods divide recordings from each subject into short windows, train classifiers on the resulting instances, and evaluate performance primarily at the instance level \cite{Medformer, DSAINet}. However, both the available disease label and the final diagnostic decision pertain to the subject as a whole. Instance classification therefore serves as a surrogate task for subject-level diagnosis, and strong performance on individual instances does not necessarily yield reliable subject-level predictions.

The underlying issue is that instance-level learning is weakly supervised because individual instances lack independent diagnostic annotations \cite{Attention}. Conventional pipelines nevertheless assign the subject label to all instances \cite{Medformer} and train the model as if they had been independently annotated. This assumes that the subject label provides an equally reliable training target for each instance, although their diagnostic relevance may differ \cite{sadatnejad2019eeg,ELMs}. The model is therefore optimized to predict the subject diagnosis from each instance, rather than to learn how they jointly support the subject-level decision. Majority voting or averaging \cite{Majority-Vote} can aggregate these predictions at inference time, but does not alter the inherited supervision used to train the encoder and classifier \cite{MTDNet}.

\begin{figure}[t]
    \centering
    \includegraphics[width=\linewidth]{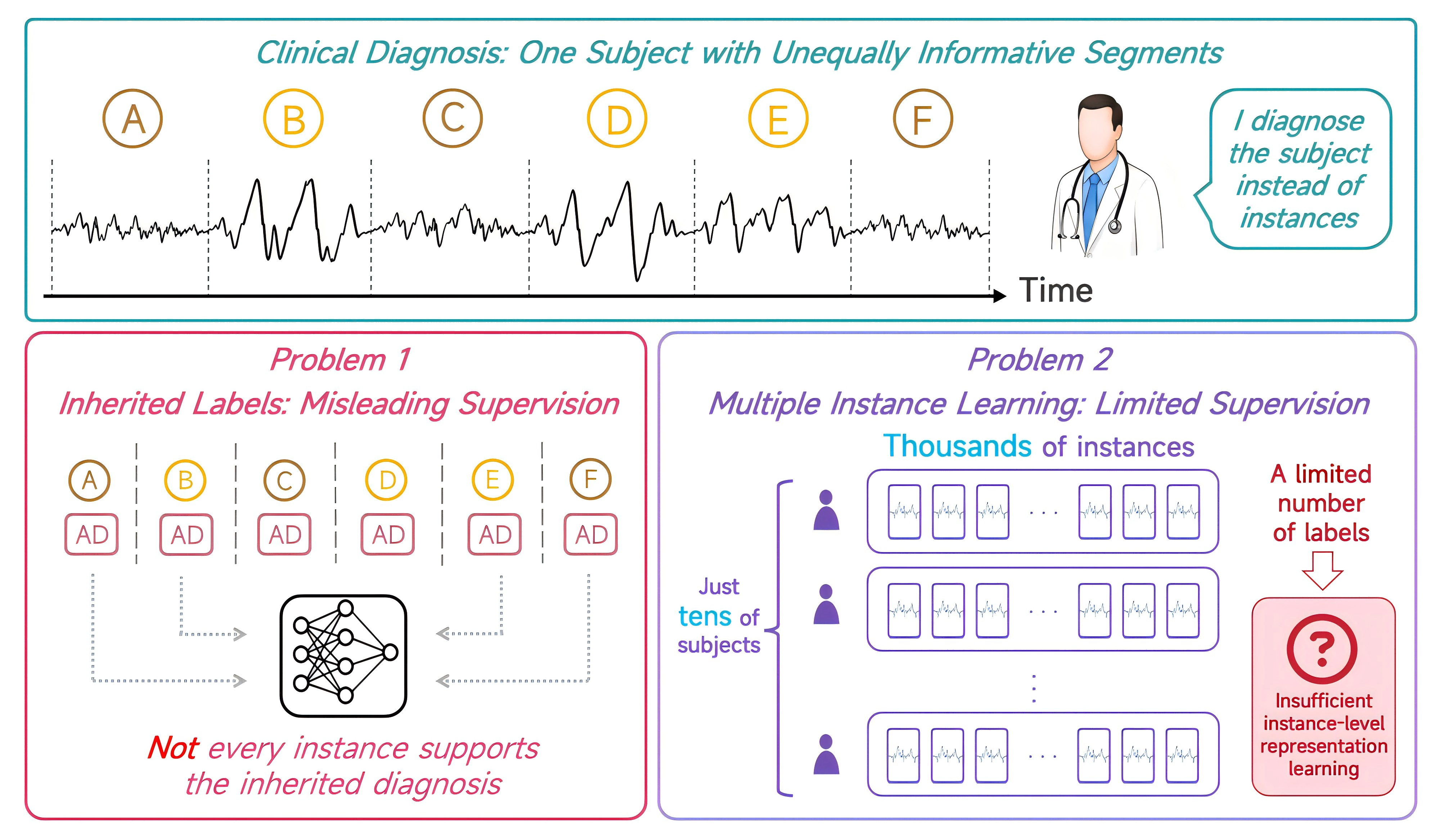}
    \caption{Motivation for BridgeMIL. EEG segments from the same subject may differ in diagnostic relevance. Inherited-label training provides dense but potentially misleading instance-level supervision, whereas MIL preserves the correct supervision level but must learn representations and aggregation from few labeled subjects.}
    \label{fig:motivation}
\end{figure}

Multiple instance learning (MIL) offers a natural formulation for subject-level EEG diagnosis by representing each subject as a bag of EEG instances and using only the subject label to supervise the bag prediction \cite{Attention, sadatnejad2019eeg}. This avoids assigning diagnosis labels to individual instances and aligns model optimization with the final subject-level decision. However, clinical EEG datasets can contain orders of magnitude more instances than labeled subjects \cite{MedGNN}. Because all instances from the same subject share one bag-level label, MIL receives only one labeled bag per subject, regardless of how many instances the bag contains. End-to-end MIL must therefore use a limited number of subject-level labels to jointly train the EEG encoder and aggregator, which may provide insufficient supervision for learning informative instance representations \cite{ItS2CLR, li2021dual}. As illustrated in Figure~\ref{fig:motivation}, subject-level EEG diagnosis presents a two-sided supervision problem: inherited-label training provides dense but potentially unreliable instance-level supervision, whereas end-to-end MIL relies on correctly aligned but limited bag-level supervision.

To address both limitations, we propose \textbf{BridgeMIL}, a two-stage framework that decouples instance representation learning from subject-level diagnostic supervision and links the stages through encoder transfer and feature retention. In Stage 1, the EEG encoder is pretrained without inherited diagnosis labels by exploiting two forms of within-subject structure. Temporally nearby instances are aligned to capture local temporal consistency, while independently sampled sub-bags from the same subject are aligned to capture information shared across different portions of the recording. An invariance objective enforces both alignments, while variance and covariance regularization prevent representation collapse and reduce redundancy without negative pairs. In Stage 2, the pretrained encoder is fine-tuned within an attention-based MIL model, where diagnostic supervision is applied only to the subject-level prediction. Because Stage 2 has access to only a limited number of labeled bags, a feature-retention objective anchors the fine-tuned encoder to the Stage 1 representation space. BridgeMIL therefore learns the encoder from abundant EEG instances without inherited labels and reserves subject labels for subject-level aggregation and prediction.

Our contributions can be summarized as follows:
\begin{itemize}
    \item We characterize a two-sided supervision problem in subject-level EEG diagnosis. Inherited-label training assigns the subject diagnosis to every instance, creating dense but potentially unreliable instance-level targets, whereas end-to-end MIL applies supervision at the correct subject level but must learn both the encoder and aggregator from only a small number of labeled subjects.
    \item We propose BridgeMIL, a two-stage framework that pretrains the EEG encoder without inherited labels through nearby-instance and within-subject sub-bag alignment with variance and covariance regularization, and then transfers the pretrained encoder to an attention-based MIL model for subject-level prediction, with feature retention limiting representation drift during fine-tuning.
    \item We evaluate BridgeMIL on three public EEG disease datasets with five representative backbones. It achieves the highest mean accuracy in 14 of 15 dataset--backbone settings. Further analyses reveal heterogeneous inherited-label reliability across instances, greater sensitivity to subject scarcity than to instance scarcity, and more structured representations with clearer subject-wise organization and diagnostic-class separability.
\end{itemize}


\section{Related Work}
\subsection{EEG-Based Disease Diagnosis}
EEG-based disease diagnosis ultimately seeks to infer a subject's disease status from recordings comprising multiple short EEG instances. Recent studies have enhanced instance representations through multi-granularity patching \cite{Medformer}, token sparsification \cite{MedSpaformer}, core-token aggregation and redistribution \cite{CoTAR}, and multi-scale temporal modeling \cite{DSAINet}. In this setting, some approaches derive subject-level diagnoses by aggregating predictions from the same subject through majority voting or score averaging \cite{chen2024multi,MTDNet}. However, their classifiers are still trained on individual instances using labels inherited from the corresponding subject. The subject-level prediction is formed only at inference by aggregating instance predictions. Better instance-level performance does not necessarily lead to better subject-level performance, because the latter depends on the distribution of predictions within each subject \cite{MTDNet}. More importantly, disease-related patterns may not be equally evident throughout a recording. Some instances may therefore contain limited diagnostic information despite sharing the same subject label. Improving instance representations or aggregating predictions at inference does not address this limitation, because the encoder is still trained to predict the subject diagnosis from every instance.

\begin{figure*}[!t]
    \centering
    \includegraphics[width=\linewidth]{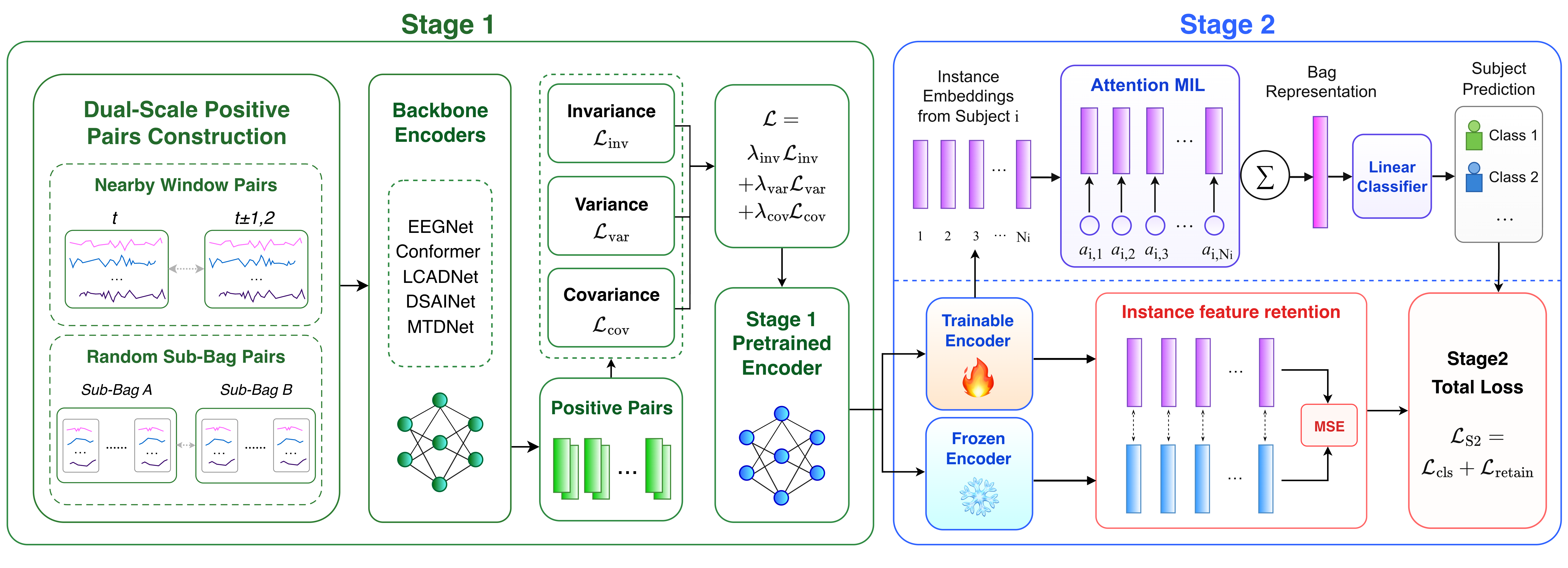}
    \caption{Overview of BridgeMIL. Stage~1 learns the EEG encoder
    without inherited instance labels by aligning temporally nearby instances
    and independently sampled sub-bags from the same subject. Variance and covariance regularization prevent representation collapse and reduce redundancy
    without negative pairs. Stage~2 transfers the pretrained encoder to an
    attention-based MIL model, while feature retention limits representation
    drift during subject-level fine-tuning.}
    \label{fig:bridgemil}
\end{figure*}

\subsection{Multiple Instance Learning}
Multiple instance learning (MIL) offers a natural formulation for subject-level EEG classification by representing each subject as a bag of EEG instances and using only the subject label for supervision. Generic MIL methods aggregate instance representations through learned attention weights \cite{Attention} or express bag-level evidence as a sum of instance-wise contributions \cite{Additive}. Time-series extensions include MILLET \cite{MILLET} for localized explanations, TimeMIL \cite{TimeMIL} for modeling temporal dependencies, and TAIL-MIL \cite{TAILMIL} for localizing anomalies across temporal and variable dimensions. In EEG emotion recognition, EmotionMIL \cite{EmotionMIL} aggregates temporally segmented instances under an overall emotion label, while MILFNet \cite{MILFNet} uses signal-level labels for weakly supervised cross-subject recognition. Attention-based MIL has also been applied to neonatal seizure detection by aggregating channel-wise time-frequency representations using only segment-level labels \cite{larbi2025neonatal}. These studies demonstrate the utility of MIL in weakly supervised time-series and EEG tasks, but their bags typically correspond to individual sequences, trials, or recording segments rather than entire subjects. In subject-level disease diagnosis, each subject provides only one labeled bag despite containing many EEG instances. Such limited bag-level supervision may hinder the joint learning of an informative encoder and a reliable aggregator for subject-level prediction.

\subsection{Self-Supervised EEG Representation Learning}
Self-supervised learning exploits structure within EEG recordings to learn instance representations without assigning subject-level labels to individual instances. EEG-specific methods construct pretext tasks from temporal context and inter-channel dependencies \cite{banville2021uncovering,MBrain}. Other approaches exploit hierarchical relations across observations, samples, trials, and patients \cite{COMET}, or align representations and transition patterns across matched brain states \cite{State-Mamba}. Masked modeling provides another route by recovering missing portions of multivariate signals from their observed context \cite{Masked-reconstruction}, and has been extended to large-scale EEG pretraining \cite{LaBraM,CBraMod,REVE}. Together, these studies demonstrate the feasibility of learning EEG representations from signal structure and recording organization. However, how such representations can support subject-level MIL under limited bag-level supervision remains underexplored.

\section{Method}

\subsection{Overview}

\textbf{BridgeMIL} addresses the two-sided supervision problem in subject-level EEG diagnosis by decoupling instance representation learning from subject-level diagnostic learning. As illustrated in Figure~\ref{fig:bridgemil}, Stage~1 exploits abundant EEG instances through nearby-window and within-subject sub-bag alignment, with variance and covariance regularization preventing representation collapse and reducing feature redundancy. Stage~2 transfers this encoder to an attention-based MIL model, which learns how the resulting instance representations should be aggregated for subject-level prediction. Because Stage~2 is optimized from comparatively few labeled subject bags, feature retention constrains encoder drift during fine-tuning. Encoder transfer and retention therefore form the bridge between representation learning from EEG instances and diagnosis under subject-level supervision.

\subsection{Problem Formulation}

Let the training set contain $S$ subjects,
\begin{equation}
    \mathcal{D}
    =
    \left\{
    \left(\mathcal{B}_i,y_i\right)
    \right\}_{i=1}^{S},
    \qquad
    \mathcal{B}_i
    =
    \left\{
    \mathbf{x}_{i,t}
    \right\}_{t=1}^{N_i},
\end{equation}
where $\mathcal{B}_i$ is the bag of $N_i$ EEG instances from
subject $i$, $\mathbf{x}_{i,t}$ is a short multichannel EEG window,
and $y_i\in\{1,\ldots,K\}$ is the subject diagnosis. No independent
label is available for any $\mathbf{x}_{i,t}$. The goal is therefore
to learn a bag-level predictor $F(\mathcal{B}_i)$ without treating
$y_i$ as a ground-truth label for every instance in the bag.


\subsection{Stage 1: Dual-Scale Feature-Alignment without Inherited Instance Labels}


Stage~1 trains an instance encoder $f_{\theta}$ and a projection head
$q_{\phi}$ using two positive-pair
constructions. Nearby-window pairs impose local temporal consistency,
while sub-bag pairs capture information shared across independently
sampled portions of the same subject's recording.
No instance-level supervision is introduced throughout the proposed BridgeMIL.

\subsubsection{Nearby-Window Alignment}

For an anchor window $\mathbf{x}_{i,t}$, we sample a second valid
window $\mathbf{x}_{i,t'}$ from the same subject within a temporal
radius $r$, where
\begin{equation}
    0 < |t'-t| \leq r.
\end{equation}

The corresponding projected representations are
\begin{equation}
    \mathbf{z}_{i,t}
    =
    q_{\phi}
    \left(
    f_{\theta}(\mathbf{x}_{i,t})
    \right),
    \qquad
    \mathbf{z}_{i,t'}
    =
    q_{\phi}
    \left(
    f_{\theta}(\mathbf{x}_{i,t'})
    \right).
\end{equation}

Aligning these representations encourages the encoder to preserve
short-range physiological structure while remaining robust to small
temporal shifts within a recording.

\subsubsection{Within-Subject Sub-Bag Alignment}

A single EEG window may contain incomplete diagnostic evidence.
We therefore construct a second positive relation at a broader scale. For subject $i$, two index sets
$\mathcal{I}^{(1)}_i$ and $\mathcal{I}^{(2)}_i$ are sampled
independently from its bag. Each sampled sub-bag is represented by mean
pooling its projected instance representations:
\begin{equation}
    \mathbf{s}^{(v)}_i
    =
    \frac{1}{|\mathcal{I}^{(v)}_i|}
    \sum_{t\in\mathcal{I}^{(v)}_i}
    q_{\phi}
    \left(
    f_{\theta}(\mathbf{x}_{i,t})
    \right),
    \qquad
    v\in\{1,2\}.
\end{equation}
The pair
$\left(\mathbf{s}^{(1)}_i,\mathbf{s}^{(2)}_i\right)$
encourages representations sampled from different portions of the same
recording to retain their shared within-subject structure.

\subsubsection{VICReg Objective}

Both positive-pair types are optimized with the VICReg objective
\cite{vicreg}. Consider a minibatch of $M$ paired projected representations,
arranged as
$\mathbf{Z}^{(1)},\mathbf{Z}^{(2)}\in\mathbb{R}^{M\times d}$,
where $d$ is the projection dimension.
The invariance term aligns  representations:
\begin{equation}
    \mathcal{L}_{\mathrm{inv}}
    =
    \frac{1}{Md}
    \left\|
    \mathbf{Z}^{(1)}
    -
    \mathbf{Z}^{(2)}
    \right\|_{F}^{2}.
\end{equation}

For view $v$, let
$\sigma_j(\mathbf{Z}^{(v)})$ denote the batch standard deviation of
representation dimension $j$. The variance term prevents dimensional
collapse:
\begin{equation}
    \mathcal{L}_{\mathrm{var}}
    =
    \frac{1}{d}
    \sum_{j=1}^{d}
    \sum_{v=1}^{2}
    \max
    \left(
    0,
    \gamma-\sigma_j(\mathbf{Z}^{(v)})
    \right),
\end{equation}
where $\gamma$ is the target standard deviation.

Let $\mathbf{C}(\mathbf{Z}^{(v)})$ denote the empirical covariance
matrix of view $v$. The covariance term reduces redundancy between
different representation dimensions:
\begin{equation}
    \mathcal{L}_{\mathrm{cov}}
    =
    \frac{1}{d}
    \sum_{v=1}^{2}
    \sum_{p\neq q}
    \left[
    \mathbf{C}(\mathbf{Z}^{(v)})
    \right]_{pq}^{2}.
\end{equation}

The VICReg objective for one pair construction is:
\begin{equation}
    \mathcal{L}_{\mathrm{VIC}}
    =
    \lambda_{\mathrm{inv}}
    \mathcal{L}_{\mathrm{inv}}
    +
    \lambda_{\mathrm{var}}
    \mathcal{L}_{\mathrm{var}}
    +
    \lambda_{\mathrm{cov}}
    \mathcal{L}_{\mathrm{cov}},
\end{equation}
where $\lambda_{\mathrm{inv}}$, $\lambda_{\mathrm{var}}$, and
$\lambda_{\mathrm{cov}}$ control the three loss terms.

We apply this objective separately to the nearby-window pairs and the
within-subject sub-bag pairs, and average the resulting losses:
\begin{equation}
    \mathcal{L}_{\mathrm{S1}}
    =
    \frac{1}{2}
    \left(
    \mathcal{L}_{\mathrm{near}}
    +
    \mathcal{L}_{\mathrm{subbag}}
    \right).
\end{equation}

The variance and covariance terms prevent the representation collapse in absence of negative
pairs, while the two alignment scales jointly exploit the large number
of available EEG instances. After Stage 1, the projection
head $q_{\phi}$ is discarded and only the pretrained encoder is
transferred to Stage~2.

\subsection{Stage 2: Subject-Level Attention MIL}


Let $\theta_0$ denote the encoder parameters obtained from Stage~1.
The Stage~2 encoder $f_{\theta}$ is initialized as
$\theta\leftarrow\theta_0$ and produces an embedding for every
instance:

\begin{equation}
    \mathbf{h}_{i,t}
    =
    f_{\theta}(\mathbf{x}_{i,t}).
\end{equation}

A fixed sinusoidal positional encoding $\mathbf{p}_t$ is added to each
instance embedding, followed by dropout:
\begin{equation}
    \widetilde{\mathbf{h}}_{i,t}
    =
    \operatorname{Dropout}
    \left(
    \mathbf{h}_{i,t}+\mathbf{p}_t
    \right).
\end{equation}

An attention network assigns an independent relevance gate to each
instance:
\begin{equation}
    a_{i,t}
    =
    \mathrm{sigmoid}
    \left(
    \mathbf{w}_{a}^{\top}
    \tanh
    \left(
    \mathbf{W}_{a}
    \widetilde{\mathbf{h}}_{i,t}
    +
    \mathbf{b}_{a}
    \right)
    +
    b_a
    \right),
\end{equation}
where $\mathbf{W}_{a}$, $\mathbf{b}_{a}$,
$\mathbf{w}_{a}$, and $b_a$ are trainable attention parameters.

The subject representation is obtained by averaging the gated instance
features:
\begin{equation}
    \mathbf{b}_i
    =
    \frac{1}{N_i}
    \sum_{t=1}^{N_i}
    a_{i,t}
    \widetilde{\mathbf{h}}_{i,t}.
\end{equation}

A subject classifier $c_{\psi}$ maps the aggregated representation to
subject-level logits:
\begin{equation}
    \mathbf{o}_i
    =
    c_{\psi}(\mathbf{b}_i).
\end{equation}

The gates are not softmax-normalized and therefore need not sum to one.
Instead, they independently modulate the contribution of each instance
before bag-level averaging. Crucially, diagnostic supervision is
applied only to the subject prediction. For a minibatch containing
$S_b$ subjects, the classification objective is
\begin{equation}
    \mathcal{L}_{\mathrm{cls}}
    =
    -
    \frac{1}{S_b}
    \sum_{i=1}^{S_b}
    \log
    \frac{
        \exp(o_{i,y_i})
    }{
        \sum_{k=1}^{K}\exp(o_{i,k})
    }.
\end{equation}


\subsection{Feature Retention Across the Bridge}

Although Stage~2 applies supervision at the correct level, it learns from only a limited number of labeled subject bags. Unconstrained
fine-tuning may therefore overwrite the representation learned from the
much larger set of EEG instances.

To limit this drift, we preserve a frozen copy of the Stage~1 encoder,
denoted by $f_{\theta_0}$, and compare its features with those produced
by the current Stage~2 encoder. For a sampled set $\mathcal{Q}$ of valid training instances, the
instance-level retention objective is
\begin{equation}
    \mathcal{L}_{\mathrm{ret}}
    =
    \frac{1}{|\mathcal{Q}|d_h}
    \sum_{\mathbf{x}\in\mathcal{Q}}
    \left\|
    f_{\theta}(\mathbf{x})
    -
    \mathrm{sg}
    \left(
    f_{\theta_0}(\mathbf{x})
    \right)
    \right\|_{2}^{2},
\end{equation}
where $d_h$ is the encoder feature dimension, $\mathrm{sg}$ denotes
stop-gradient, and $\theta_0$ remains fixed throughout Stage~2.

The final Stage~2 objective combines subject-level classification and
feature retention:
\begin{equation}
    \mathcal{L}_{\mathrm{S2}}
    =
    \mathcal{L}_{\mathrm{cls}}
    +
    \lambda_{\mathrm{ret}}
    \mathcal{L}_{\mathrm{ret}},
\end{equation}
where $\lambda_{\mathrm{ret}}$ controls the strength of feature
retention during fine-tuning.

BridgeMIL therefore reserves disease labels to learn how instance
representations should be aggregated into a subject prediction.
The transferred encoder and feature-retention objective preserve the
representation learned without inherited instance labels. At
inference,
only the fine-tuned encoder, attention aggregator, and subject
classifier are retained.

\section{Experiments and Results}

\begin{table*}[t]
\centering
{\small
\setlength{\tabcolsep}{3pt}
\begin{tabular}{lcccccccc}
\toprule
\multirow{2}{*}{Backbone}
& \multirow{2}{*}[-0.5ex]{\makecell{Majority\\Vote}}
& \multicolumn{2}{c}{Generic MIL}
& \multicolumn{2}{c}{Modern MIL}
& \multicolumn{3}{c}{Two-Stage Representation Learning} \\
\cmidrule(lr){3-4}
\cmidrule(lr){5-6}
\cmidrule(lr){7-9}
&
& Additive
& Attention
& MILLET
& TimeMIL
& SupCon
& Masked Recon.
& BridgeMIL \\
\midrule

\multicolumn{9}{c}{
ADFTD (Alzheimer's Disease \& Frontotemporal Dementia,
3 Classes, 88 Subjects, 69,794 Instances)
} \\
\midrule

EEGNet
& 55.74 $\pm$ 1.55
& 58.53 $\pm$ 0.43
& 58.10 $\pm$ 1.18
& 58.03 $\pm$ 1.01
& 57.01 $\pm$ 0.94
& \underline{60.68 $\pm$ 0.69}
& 42.12 $\pm$ 1.71
& \textbf{63.40 $\pm$ 0.95} \\

Conformer
& 55.91 $\pm$ 1.05
& 54.35 $\pm$ 1.76
& 56.59 $\pm$ 1.77
& 59.34 $\pm$ 1.60
& \underline{59.78 $\pm$ 0.47}
& 56.95 $\pm$ 0.90
& 59.22 $\pm$ 1.43
& \textbf{62.51 $\pm$ 1.38} \\

LCADNet
& 51.67 $\pm$ 1.88
& 49.77 $\pm$ 0.26
& 50.25 $\pm$ 1.01
& 50.36 $\pm$ 0.83
& 52.73 $\pm$ 1.08
& \textbf{56.14 $\pm$ 0.87}
& 48.38 $\pm$ 0.80
& \underline{54.65 $\pm$ 0.81} \\

DSAINet
& 55.43 $\pm$ 2.34
& 58.70 $\pm$ 2.79
& 56.18 $\pm$ 0.64
& 57.53 $\pm$ 0.74
& 60.67 $\pm$ 1.23
& \underline{61.90 $\pm$ 0.62}
& 47.63 $\pm$ 0.89
& \textbf{64.23 $\pm$ 1.36} \\

MTDNet
& 56.45 $\pm$ 0.92
& 58.82 $\pm$ 1.32
& \underline{59.01 $\pm$ 2.02}
& 58.09 $\pm$ 2.43
& 58.74 $\pm$ 0.72
& 55.35 $\pm$ 1.95
& 41.61 $\pm$ 0.58
& \textbf{63.27 $\pm$ 1.65} \\

\midrule
\multicolumn{9}{c}{
Mumtaz2017 (Major Depressive Disorder,
2 Classes, 63 Subjects, 35,891 Instances)
} \\
\midrule

EEGNet
& 85.76 $\pm$ 1.16
& 83.38 $\pm$ 0.92
& 87.00 $\pm$ 1.97
& \underline{87.72 $\pm$ 1.71}
& 83.29 $\pm$ 1.67
& 87.57 $\pm$ 0.73
& 71.14 $\pm$ 0.71
& \textbf{91.51 $\pm$ 1.46} \\

Conformer
& 87.52 $\pm$ 1.98
& 85.06 $\pm$ 1.33
& 85.43 $\pm$ 1.66
& 88.57 $\pm$ 0.84
& 83.81 $\pm$ 2.03
& \underline{88.95 $\pm$ 0.12}
& 88.14 $\pm$ 0.55
& \textbf{92.22 $\pm$ 1.17} \\

LCADNet
& \underline{84.71 $\pm$ 1.71}
& 73.22 $\pm$ 0.84
& 78.43 $\pm$ 2.25
& 78.76 $\pm$ 2.04
& 80.59 $\pm$ 1.95
& 84.29 $\pm$ 1.10
& 78.57 $\pm$ 0.65
& \textbf{88.10 $\pm$ 1.38} \\

DSAINet
& 87.43 $\pm$ 1.20
& 86.76 $\pm$ 2.30
& 88.62 $\pm$ 2.20
& \underline{89.95 $\pm$ 0.98}
& 89.00 $\pm$ 1.18
& 88.43 $\pm$ 0.53
& 80.76 $\pm$ 0.38
& \textbf{93.39 $\pm$ 1.17} \\

MTDNet
& 86.95 $\pm$ 0.53
& 83.09 $\pm$ 1.66
& 88.67 $\pm$ 1.21
& \underline{90.67 $\pm$ 1.89}
& 88.00 $\pm$ 1.27
& 89.57 $\pm$ 0.57
& 77.14 $\pm$ 1.56
& \textbf{91.81 $\pm$ 0.91} \\

\midrule
\multicolumn{9}{c}{
Rockhill2021 (Parkinson's Disease,
2 Classes, 31 Subjects, 12,033 Instances)
} \\
\midrule

EEGNet
& 63.14 $\pm$ 0.35
& 63.52 $\pm$ 1.72
& 62.00 $\pm$ 1.18
& 61.43 $\pm$ 2.11
& 66.67 $\pm$ 2.66
& \underline{71.62 $\pm$ 1.37}
& 51.71 $\pm$ 0.23
& \textbf{77.38 $\pm$ 2.10} \\

Conformer
& 59.43 $\pm$ 1.07
& 56.00 $\pm$ 1.75
& 56.57 $\pm$ 2.70
& 54.67 $\pm$ 2.88
& 64.10 $\pm$ 3.40
& \underline{70.29 $\pm$ 0.88}
& 56.29 $\pm$ 1.69
& \textbf{78.10 $\pm$ 2.24} \\

LCADNet
& 64.67 $\pm$ 2.42
& 57.43 $\pm$ 1.34
& 56.10 $\pm$ 1.64
& 62.95 $\pm$ 1.26
& 63.71 $\pm$ 2.67
& \underline{69.24 $\pm$ 2.28}
& 57.53 $\pm$ 1.29
& \textbf{72.67 $\pm$ 1.97} \\

DSAINet
& 61.24 $\pm$ 1.50
& 61.14 $\pm$ 0.48
& 65.62 $\pm$ 1.29
& 64.00 $\pm$ 1.11
& 64.29 $\pm$ 2.20
& \underline{76.86 $\pm$ 0.93}
& 56.47 $\pm$ 2.27
& \textbf{77.71 $\pm$ 1.28} \\

MTDNet
& 64.29 $\pm$ 1.20
& 64.95 $\pm$ 1.37
& 64.38 $\pm$ 1.55
& 66.00 $\pm$ 1.15
& 64.38 $\pm$ 3.40
& \underline{66.57 $\pm$ 1.55}
& 58.57 $\pm$ 1.48
& \textbf{77.62 $\pm$ 1.88} \\

\bottomrule
\end{tabular}
}
\caption{Accuracy across three datasets (mean $\pm$ standard deviation over five seeds, \%; best in bold, second best underlined).}
\label{tab:main_results}
\end{table*}

\subsection{Experiment Setup}
We evaluate BridgeMIL for subject-level EEG disease diagnosis on three public datasets with five representative backbones and address the following research questions:
\begin{itemize}
    \item \textbf{RQ1:} How does BridgeMIL compare with majority vote, end-to-end MIL, and alternative two-stage representation-learning approaches across EEG datasets and backbone architectures? (\textbf{Performance Comparison})
    \item \textbf{RQ2:} Do inherited subject labels provide equally reliable supervision across EEG instances from the same subject? (\textbf{Inherited-Label Reliability Analysis})
    \item \textbf{RQ3:} How do subject scarcity and instance scarcity affect subject-level EEG diagnosis? (\textbf{Subject and Instance Scarcity Analysis})
    \item \textbf{RQ4:} How do BridgeMIL's Stage 1 objective and Stage 2 feature retention contribute to its learned representations and subject-level performance? (\textbf{Ablation and Representation Analysis})

\end{itemize}

\subsubsection{Datasets}
We use three public EEG disease datasets: ADFTD~\cite{ADFTD}, Mumtaz2017~\cite{Mumtaz2017}, and Rockhill2021~\cite{Rockhill2021}. For Mumtaz2017, we retain only eyes-open and eyes-closed sessions; for Rockhill2021, only medication-ON recordings. Further dataset details are provided in Appendix A.

\subsubsection{Preprocessing}
All EEG recordings are band-pass filtered and downsampled to 200 Hz before being segmented into 1-s instances. The passbands are 0.5-45 Hz for ADFTD, 0.3-75 Hz for Mumtaz2017, and 0.5-40 Hz for Rockhill2021. For Mumtaz2017, a 50 Hz notch filter is additionally applied to suppress power-line noise. Rockhill2021 is segmented with 50\% overlap between adjacent instances, whereas ADFTD and Mumtaz2017 are segmented without overlap. Within each split, channel-wise z-score statistics were estimated from the training set and applied unchanged to the training, validation, and test sets.

\subsubsection{Evaluation Protocol}
We adopt subject-level stratified $k$-fold cross-validation, with all instances from a given subject assigned to the same fold. For each round, one fold is held out for testing, another for validation, and the remaining $k-2$ folds are used for training. We set $k=10$ for all datasets except Rockhill2021, for which $k=5$ is used to account for its smaller number of subjects. Only the training fold is used throughout the two stages of BridgeMIL.

\subsubsection{Backbones}
We use five representative backbones: EEGNet~\cite{EEGNet}, Conformer~\cite{Conformer}, LCADNet~\cite{LCADNet}, DSAINet~\cite{DSAINet} and MTDNet~\cite{MTDNet}.

\subsubsection{Baselines}
Baselines include majority voting~\cite{Majority-Vote}, generic MIL (attention-based MIL~\cite{Attention} and Additive MIL~\cite{Additive}), modern MIL (MILLET~\cite{MILLET} and TimeMIL~\cite{TimeMIL}), and two-stage alternatives that replace Stage~1 with SupCon~\cite{SupCon} or masked reconstruction~\cite{Masked-reconstruction} before the same attention-based MIL stage. Detailed backbone and method implementations are provided in Appendix D.


\begin{figure}[t]
    \centering
    \includegraphics[width=0.48\linewidth]{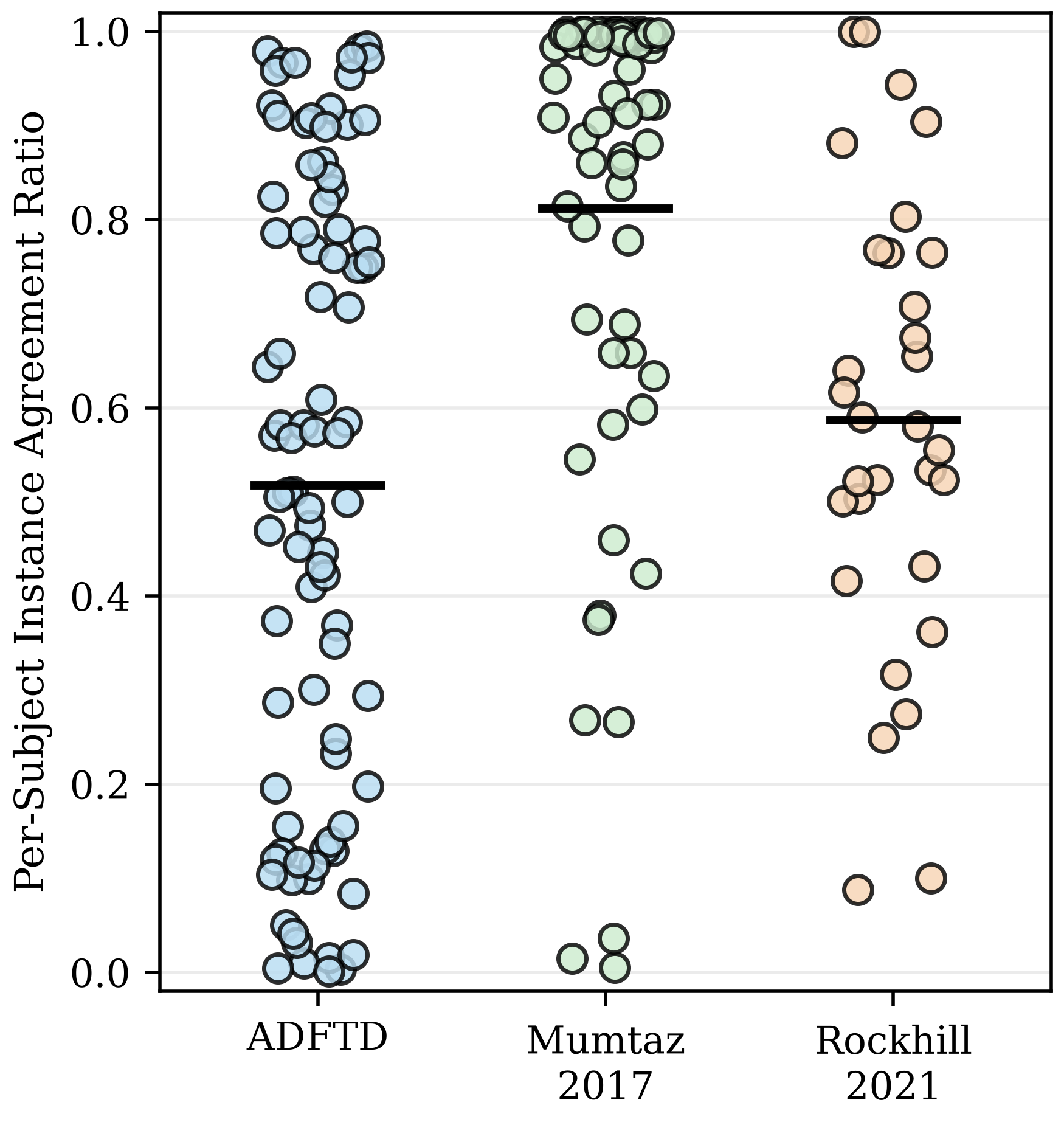}
    \includegraphics[width=0.48\linewidth]{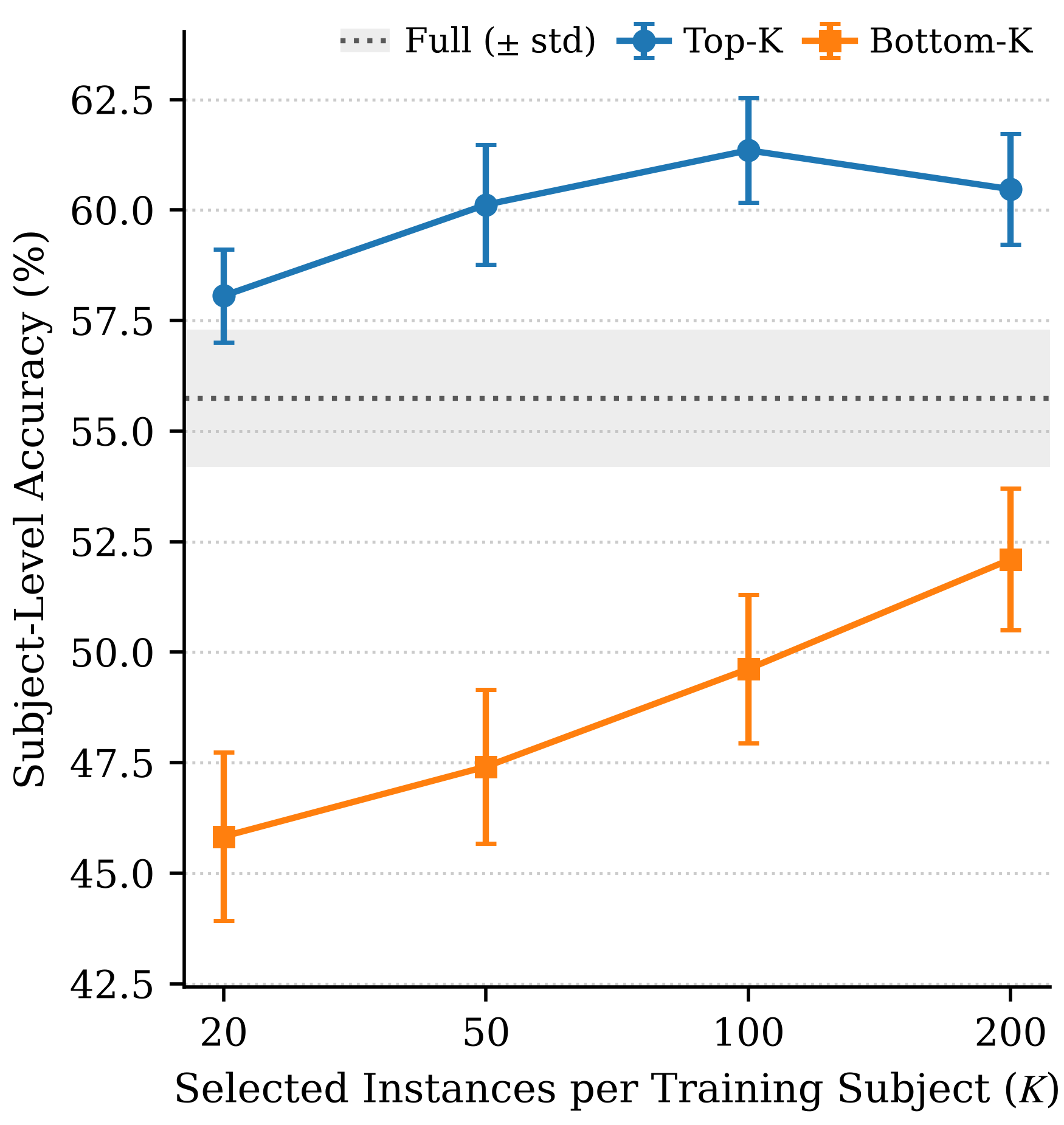}
    \caption{Inherited-label reliability analysis. \textbf{Left:} Per-subject instance agreement ratios across three datasets, with horizontal bars indicating the mean ratio for each dataset. \textbf{Right:} Subject-level accuracy on ADFTD using Top-K, Bottom-K, or all training instances. Results are mean $\pm$ standard deviation over five seeds.}
    \label{fig:label_reliability}
\end{figure}

\subsection{RQ1: Performance Comparison}



Table~\ref{tab:main_results} compares inherited-label training with majority voting, end-to-end MIL, alternative two-stage representation learning, and BridgeMIL. Across three datasets and five backbones, BridgeMIL achieves the highest mean accuracy in 14 of 15 dataset--backbone settings and outperforms majority voting and every end-to-end MIL baseline in all 15. Its overall mean accuracy is 76.57\%, exceeding the strongest baseline, SupCon (72.29\%), by 4.28 percentage points. The only exception is LCADNet on ADFTD, where SupCon obtains 56.14\% compared with 54.65\% for BridgeMIL. 

These comparisons suggest that addressing only one aspect of the supervision problem is insufficient. Across all 15 settings, BridgeMIL exceeds majority voting by an average of 8.55 points, indicating that inference-time aggregation does not resolve the supervision mismatch introduced by inherited-label training. It also outperforms the strongest end-to-end MIL method in each setting by an average of 6.28 points, suggesting that limited subject-level supervision may be insufficient to jointly learn an informative encoder and reliable aggregator. Compared with the stronger of SupCon and masked reconstruction in each setting, BridgeMIL achieves an average margin of 4.13 points, indicating that its gains cannot be explained by two-stage training alone. Overall, these results support learning instance representations without inherited diagnosis labels before subject-level aggregation and prediction.

\begin{figure}[t]
    \centering
    \includegraphics[width=0.48\linewidth]{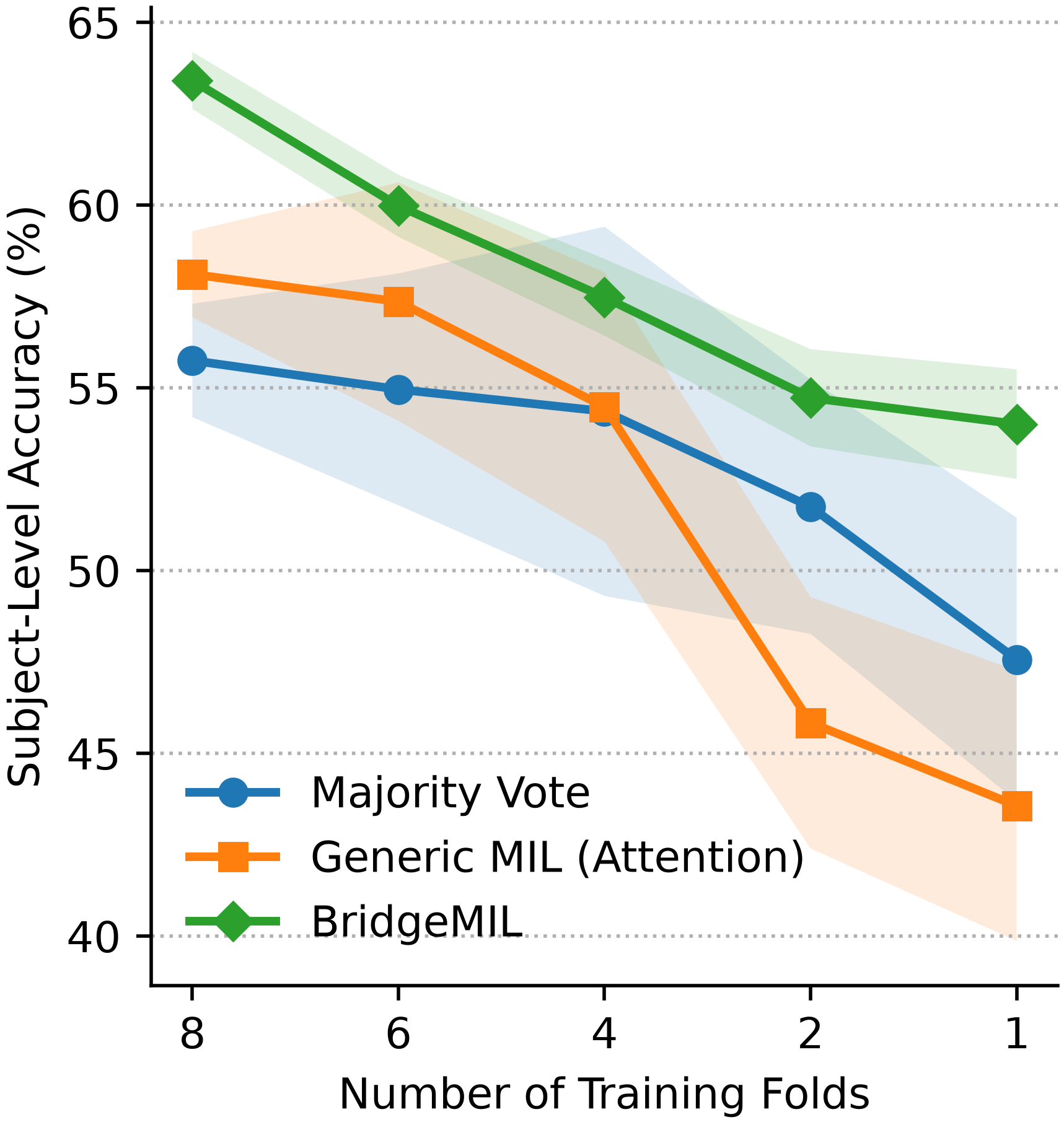}
    \includegraphics[width=0.48\linewidth]{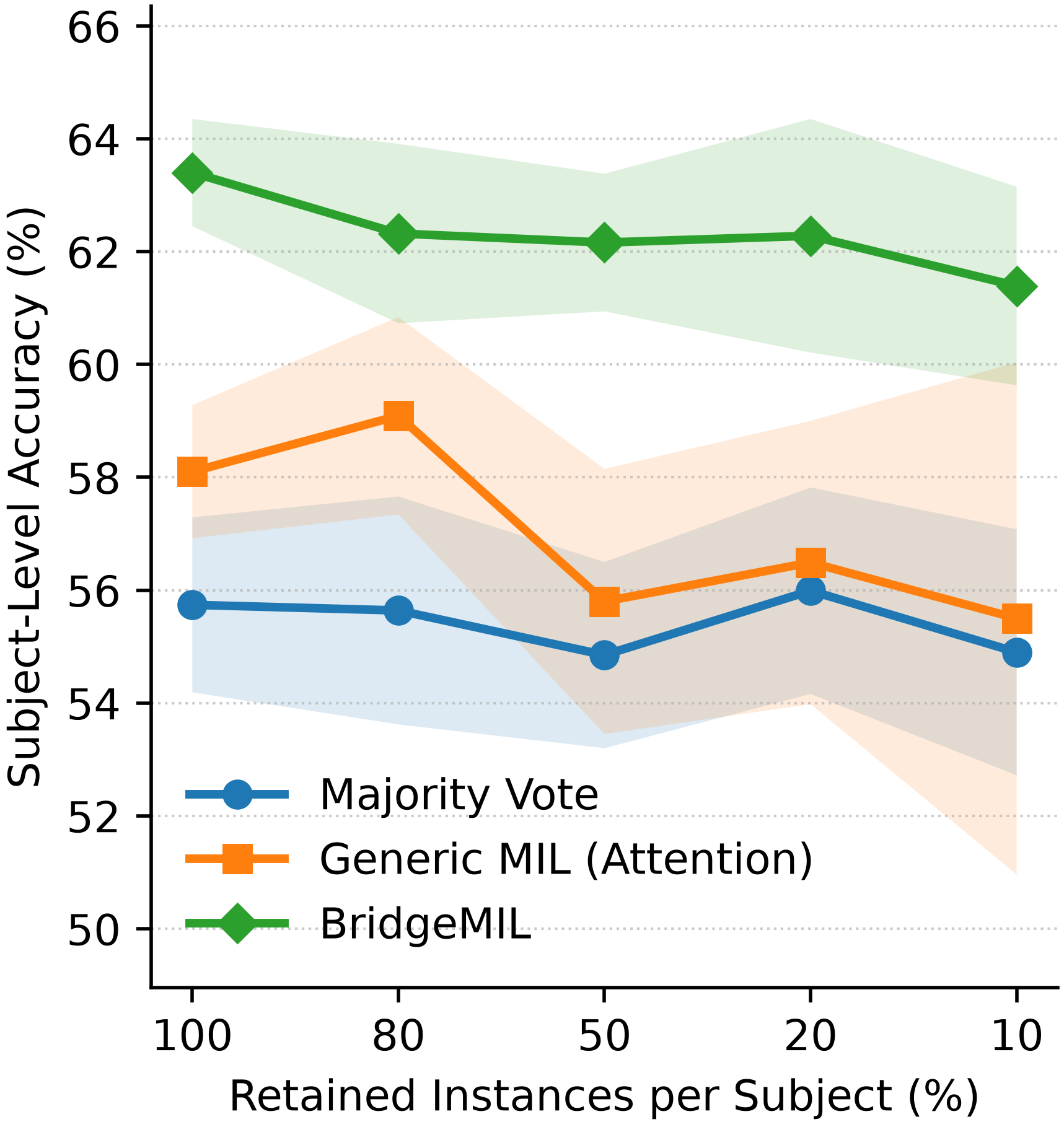}
    \caption{Subject and instance scarcity analysis on ADFTD with EEGNet. \textbf{Left:} Subject-level accuracy as the number of training folds decreases, with all instances retained. \textbf{Right:} Subject-level accuracy as the retained instances per subject decrease, with all training subjects retained. Lines and shaded regions show the mean and standard deviation over five seeds.}
    \label{fig:subject_instance_scarcity}
\end{figure}

\subsection{RQ2: Inherited-Label Reliability Analysis}

Inherited-label training assigns the subject diagnosis to every EEG instance, implicitly treating it as equally reliable supervision for all instances. Using the EEGNet majority-vote method on the three datasets, we define each held-out subject's instance agreement ratio as the proportion of instance predictions matching the subject label. As shown in the left panel of Figure~\ref{fig:label_reliability}, a substantial fraction of instances from many subjects receive predictions that disagree with the inherited subject label, suggesting that inherited labels are not uniformly reliable as instance-level supervision.

To test whether inherited labels are more reliable for some instances than others, we compare EEGNet trained on Top-K and Bottom-K instances ranked by BridgeMIL attention weights. On ADFTD, we use subject-level 10-fold cross-validation to obtain attention weights for each subject only when that subject is held out. For K = 20, 50, 100, and 200, new EEGNet classifiers are trained with inherited labels using only the selected training instances; validation and test bags remain complete, and subject-level predictions are obtained by majority voting. As shown in the right panel of Figure~\ref{fig:label_reliability}, Top-K yields higher mean accuracy than both Bottom-K and the all-instance condition at every K, while Bottom-K remains below the all-instance result. At K = 100, Top-K reaches the highest mean accuracy of 61.35\%, compared with 49.62\% for Bottom-K and 55.74\% using all training instances. This ordering is not explained by training-set size: Top-K and Bottom-K use the same number of instances, and Top-K also surpasses the larger all-instance condition. Together with the agreement ratio analysis, these results indicate that inherited labels are not equally reliable when used as instance-level supervision.

\begin{figure*}[t]
    \centering
    \includegraphics[width=\linewidth]{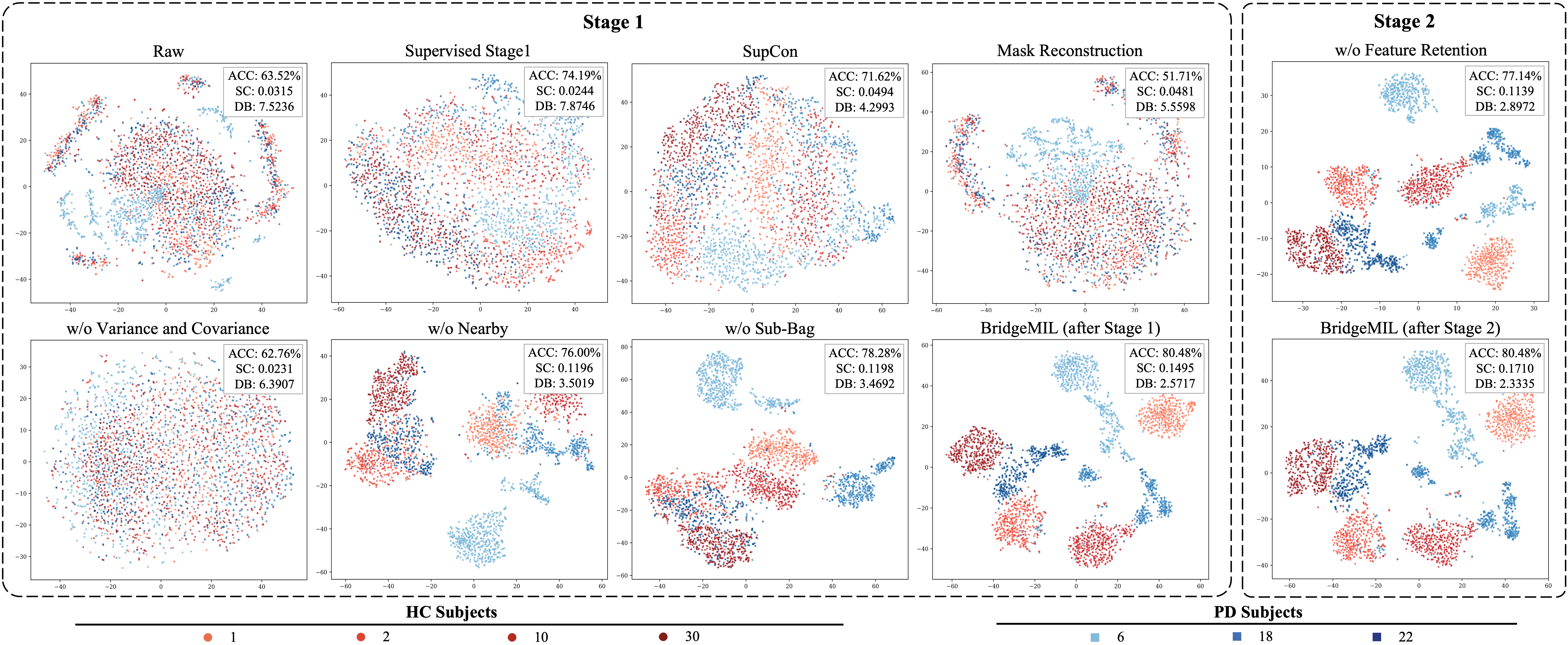}
    \caption{Ablation and representation analysis on Rockhill2021 with EEGNet. Stage 1 compares raw windows, pretraining alternatives, and component ablations, while Stage 2 compares fine-tuning with and without feature retention. Points represent EEG instances. Colors denote classes, and shades identify subjects within each class. ACC reports seed-0 K-fold subject-level accuracy. t-SNE, SC, and DB use the seed-0 fold-1 test set. Higher SC and lower DB indicate better class separation.}
    \label{fig:tsne}
\end{figure*}

\subsection{RQ3: Subject and Instance Scarcity Analysis}
Subject-level EEG diagnosis may be limited by both the number of labeled subjects and the number of instances available per subject. To separate these effects, we compare majority voting, end-to-end attention MIL, and BridgeMIL on ADFTD with EEGNet under two controlled settings. For subject scarcity, each seed uses either all ten folds or a random subset of 8, 6, 4, or 3 folds. One fold is used for validation and another for testing, yielding 8, 6, 4, 2, or 1 training folds, with all instances retained. For instance scarcity, the subject splits remain fixed, and we retain either all instances or the final 80\%, 50\%, 20\%, or 10\% of each subject's temporally ordered instances in the training, validation, and test sets. All methods use the same folds and retained instances within each condition, and results are averaged over five seeds.

As shown in the left panel of Figure~\ref{fig:subject_instance_scarcity}, reducing the number of training folds from 8 to 1 lowers accuracy by 8.18 points for majority voting, 14.54 points for attention MIL, and 9.40 points for BridgeMIL. By contrast, reducing the retained instances per subject from 100\% to 10\% produces much smaller decreases of 0.84, 2.60, and 2.01 points, respectively. This consistent pattern across all three methods indicates that performance is more sensitive to reductions in labeled subjects than in instances per subject. Attention MIL shows the largest degradation under subject scarcity, while BridgeMIL maintains the highest accuracy across both settings. Together, these results identify labeled-subject scarcity as the more critical constraint in this setting, consistent with prior EEG scaling results showing a stronger effect of participant count in diagnostic tasks \cite{bomatter2026limited}.

\subsection{RQ4: Ablation and Representation Analysis}
We analyze how BridgeMIL's Stage 1 objective and Stage 2 feature retention contribute to its representations and subject-level performance on Rockhill2021 with EEGNet. Raw windows provide a no-pretraining reference. Supervised Stage 1 training with inherited labels, SupCon, and masked reconstruction replace the proposed objective and use the same attention-based MIL stage, testing whether gains arise from pretraining alone. We remove variance and covariance regularization, nearby-window alignment, or sub-bag alignment, and compare Stage 2 with and without feature retention.

Figure~\ref{fig:tsne} reports the seed-0 K-fold subject-level accuracy. The t-SNE plots use representations from the seed-0 fold-1 test set, with colors denoting classes and shades denoting subjects. The Silhouette Coefficient (SC) and Davies--Bouldin index (DB) are computed in the original feature space using diagnostic classes as clusters. Higher SC and lower DB indicate better class separation.

With the full Stage 1 objective, accuracy reaches 80.48\%, compared with 63.52\% for raw windows, 74.19\% for supervised Stage 1, 71.62\% for SupCon, and 51.71\% for masked reconstruction. It also achieves the highest SC of 0.1495 and the lowest DB of 2.5717 among the Stage 1 variants. Removing nearby-window or sub-bag alignment reduces accuracy to 76.00\% and 78.28\%, respectively. Removing variance and covariance regularization causes the largest component-level degradation, lowering accuracy to 62.76\%, SC to 0.0231, and increasing DB to 6.3907. Consistent with these results, the t-SNE plots show clearer class separation for the full objective than for the raw reference, alternative Stage 1 objectives, and component ablations.

Feature retention raises accuracy from 77.14\% to 80.48\%, SC from 0.1139 to 0.1710, and lowers DB from 2.8972 to 2.3335; t-SNE likewise shows clearer class separation and more compact representations. The consistent changes in accuracy, SC, and DB indicate that the performance gains reflect improved representation quality rather than merely a more favorable classifier fit. Taken together, Stage~1 establishes class-discriminative instance representations, while feature retention preserves this structure during optimization with scarce subject-level labels. Detailed five-seed ablation results across all three datasets are provided in Appendix E.

\section{Conclusion}
We formulate subject-level EEG disease diagnosis as a two-sided supervision problem and propose BridgeMIL to decouple instance representation learning from subject-level supervision. Across three datasets and five backbones, BridgeMIL outperforms majority voting and end-to-end MIL baselines in all 15 settings and achieves the highest accuracy in 14 of 15. Further analyses show that inherited-label reliability varies across instances, subject scarcity is more limiting than instance scarcity, and the complete Stage 1 objective with feature retention improves subject-level performance and representations. These findings support learning from abundant EEG instances without inherited instance labels while reserving supervision for subject-level prediction.

\section*{Acknowledgments}
This work was supported in part by the National Natural Science Foundation of China (Nos. 2025ZD0215701, T2341003, and U2336214), the Beijing Natural Science Foundation (No. L257019), the General Program of the Guangdong Natural Science Foundation (No. 2026A1515010121), and the Shenzhen Science and Technology Innovation Committee (No. RCBS20231211090748082).

\bibliography{aaai2027}

\begin{thebibliography}{38}
\providecommand{\natexlab}[1]{#1}

\bibitem[{Banville et~al.(2021)Banville, Chehab, Hyv{\"a}rinen, Engemann, and Gramfort}]{banville2021uncovering}
Banville, H.; Chehab, O.; Hyv{\"a}rinen, A.; Engemann, D.-A.; and Gramfort, A. 2021.
\newblock {Uncovering the structure of clinical EEG signals with self-supervised learning}.
\newblock \emph{Journal of Neural Engineering}, 18(4): 046020.

\bibitem[{Bardes, Ponce, and Lecun(2022)}]{vicreg}
Bardes, A.; Ponce, J.; and Lecun, Y. 2022.
\newblock {VICReg: Variance-Invariance-Covariance Regularization For Self-Supervised Learning}.
\newblock In \emph{ICLR 2022-International Conference on Learning Representations}.

\bibitem[{Bomatter and Gouk(2026)}]{bomatter2026limited}
Bomatter, P.; and Gouk, H. 2026.
\newblock {Is limited participant diversity impeding eeg-based machine learning?}
\newblock \emph{Advances in Neural Information Processing Systems}, 38: 117447--117474.

\bibitem[{Cai et~al.(2023)Cai, Chen, Yang, Liu, and Li}]{MBrain}
Cai, D.; Chen, J.; Yang, Y.; Liu, T.; and Li, Y. 2023.
\newblock {Mbrain: A multi-channel self-supervised learning framework for brain signals}.
\newblock In \emph{Proceedings of the 29th ACM SIGKDD Conference on Knowledge Discovery and Data Mining}, 130--141.

\bibitem[{Chen et~al.(2024{\natexlab{a}})Chen, Zhang, Yang, Peng, Xie, Lv, and Hou}]{chen2024multi}
Chen, S.; Zhang, C.; Yang, H.; Peng, L.; Xie, H.; Lv, Z.; and Hou, Z.-G. 2024{\natexlab{a}}.
\newblock {A multi-modal classification method for early diagnosis of mild cognitive impairment and Alzheimer’s disease using three paradigms with various task difficulties}.
\newblock \emph{IEEE Transactions on Neural Systems and Rehabilitation Engineering}, 32: 1477--1486.

\bibitem[{Chen et~al.(2024{\natexlab{b}})Chen, Qiu, Zhu, Li, Wang, Sotiras, Wang, and Razi}]{TimeMIL}
Chen, X.; Qiu, P.; Zhu, W.; Li, H.; Wang, H.; Sotiras, A.; Wang, Y.; and Razi, A. 2024{\natexlab{b}}.
\newblock {TimeMIL: advancing multivariate time series classification via a time-aware multiple instance learning}.
\newblock In \emph{Proceedings of the 41st International Conference on Machine Learning}, 7190--7206.

\bibitem[{Early et~al.(2024)Early, Cheung, Cutajar, Xie, Kandola, and Twomey}]{MILLET}
Early, J.; Cheung, G.; Cutajar, K.; Xie, H.; Kandola, J.; and Twomey, N. 2024.
\newblock {Inherently interpretable time series classification via multiple instance learning}.
\newblock In \emph{International Conference on Learning Representations}, volume 2024, 56800--56828.

\bibitem[{El~Ouahidi et~al.(2026)El~Ouahidi, Lys, Th{\"o}lke, Farrugia, Pasdeloup, Gripon, Jerbi, and Lioi}]{REVE}
El~Ouahidi, Y.; Lys, J.; Th{\"o}lke, P.; Farrugia, N.; Pasdeloup, B.; Gripon, V.; Jerbi, K.; and Lioi, G. 2026.
\newblock {REVE: A foundation model for EEG-adapting to any setup with large-scale pretraining on 25,000 subjects}.
\newblock \emph{Advances in Neural Information Processing Systems}, 38: 22541--22577.

\bibitem[{Fan et~al.(2025)Fan, Fei, Guo, Yi, Song, Xiang, Ye, and Li}]{MedGNN}
Fan, W.; Fei, J.; Guo, D.; Yi, K.; Song, X.; Xiang, H.; Ye, H.; and Li, M. 2025.
\newblock {Towards multi-resolution spatiotemporal graph learning for medical time series classification}.
\newblock In \emph{Proceedings of the ACM on Web Conference 2025}, 5054--5064.

\bibitem[{Gijsen and Ritter(2025)}]{ELMs}
Gijsen, S.; and Ritter, K. 2025.
\newblock {EEG-Language Pretraining for Highly Label-Efficient Clinical Phenotyping}.
\newblock In \emph{International Conference on Machine Learning}, 19480--19504. PMLR.

\bibitem[{Ilse, Tomczak, and Welling(2018)}]{Attention}
Ilse, M.; Tomczak, J.; and Welling, M. 2018.
\newblock {Attention-based deep multiple instance learning}.
\newblock In \emph{International conference on machine learning}, 2127--2136. PMLR.

\bibitem[{Jang and Kwon(2025)}]{TAILMIL}
Jang, J.; and Kwon, H.-Y. 2025.
\newblock {TAIL-MIL: time-aware and instance-learnable multiple instance learning for multivariate time series anomaly detection}.
\newblock In \emph{Proceedings of the AAAI Conference on Artificial Intelligence}, volume~39, 17582--17589.

\bibitem[{Javed et~al.(2022)Javed, Juyal, Padigela, Taylor-Weiner, Yu, and Prakash}]{Additive}
Javed, S.~A.; Juyal, D.; Padigela, H.; Taylor-Weiner, A.; Yu, L.; and Prakash, A. 2022.
\newblock {Additive mil: Intrinsically interpretable multiple instance learning for pathology}.
\newblock \emph{Advances in Neural Information Processing Systems}, 35: 20689--20702.

\bibitem[{Jiang, Zhao, and Lu(2024)}]{LaBraM}
Jiang, W.-B.; Zhao, L.; and Lu, B.-L. 2024.
\newblock {Large brain model for learning generic representations with tremendous EEG data in BCI}.
\newblock In \emph{International Conference on Learning Representations}, volume 2024, 16405--16426.

\bibitem[{Jin et~al.(2026)Jin, Kim, Na, and Kim}]{MILFNet}
Jin, C.; Kim, H.; Na, Y.; and Kim, S.-E. 2026.
\newblock {MILFNet: Multiple Instance Learning-Based FastSlow Network With Multitask Autoencoder for Cross-Subject EEG Emotion Recognition}.
\newblock \emph{IEEE Transactions on Instrumentation and Measurement}.

\bibitem[{Kachare et~al.(2024)Kachare, Puri, Sangle, Al-Shourbaji, Jabbari, Kirner, Alameen, Migdady, and Abualigah}]{LCADNet}
Kachare, P.; Puri, D.; Sangle, S.~B.; Al-Shourbaji, I.; Jabbari, A.; Kirner, R.; Alameen, A.; Migdady, H.; and Abualigah, L. 2024.
\newblock {LCADNet: a novel light CNN architecture for EEG-based Alzheimer disease detection}.
\newblock \emph{Physical and Engineering Sciences in Medicine}, 47(3): 1037--1050.

\bibitem[{Khosla et~al.(2020)Khosla, Teterwak, Wang, Sarna, Tian, Isola, Maschinot, Liu, and Krishnan}]{SupCon}
Khosla, P.; Teterwak, P.; Wang, C.; Sarna, A.; Tian, Y.; Isola, P.; Maschinot, A.; Liu, C.; and Krishnan, D. 2020.
\newblock {Supervised contrastive learning}.
\newblock \emph{Advances in neural information processing systems}, 33: 18661--18673.

\bibitem[{Larbi et~al.(2025)Larbi, Abed, Cardoso, and Ouahabi}]{larbi2025neonatal}
Larbi, A.; Abed, M.; Cardoso, J.~S.; and Ouahabi, A. 2025.
\newblock {Neonatal EEG classification using a compact support separable kernel time--frequency distribution and attention-based CNN}.
\newblock \emph{Biomedical Signal Processing and Control}, 110: 108072.

\bibitem[{Lawhern et~al.(2018)Lawhern, Solon, Waytowich, Gordon, Hung, and Lance}]{EEGNet}
Lawhern, V.~J.; Solon, A.~J.; Waytowich, N.~R.; Gordon, S.~M.; Hung, C.~P.; and Lance, B.~J. 2018.
\newblock {EEGNet: a compact convolutional neural network for EEG-based brain--computer interfaces}.
\newblock \emph{Journal of neural engineering}, 15(5): 056013.

\bibitem[{Li, Li, and Eliceiri(2021)}]{li2021dual}
Li, B.; Li, Y.; and Eliceiri, K.~W. 2021.
\newblock {Dual-stream multiple instance learning network for whole slide image classification with self-supervised contrastive learning}.
\newblock In \emph{Proceedings of the IEEE/CVF conference on computer vision and pattern recognition}, 14318--14328.

\bibitem[{Liu et~al.(2023)Liu, Zhu, Shen, Liu, Razavian, Geras, and Fernandez-Granda}]{ItS2CLR}
Liu, K.; Zhu, W.; Shen, Y.; Liu, S.; Razavian, N.; Geras, K.~J.; and Fernandez-Granda, C. 2023.
\newblock {Multiple instance learning via iterative self-paced supervised contrastive learning}.
\newblock In \emph{Proceedings of the IEEE/CVF conference on computer vision and pattern recognition}, 3355--3365.

\bibitem[{Ma et~al.(2026)Ma, Li, Qiu, Li, Meng, Zhang, Liu, Shen, and Song}]{DSAINet}
Ma, Z.; Li, Z.; Qiu, Z.; Li, J.; Meng, L.; Zhang, X.; Liu, Y.; Shen, X.; and Song, S. 2026.
\newblock {DSAINet: An Efficient Dual-Scale Attentive Interaction Network for General EEG Decoding}.
\newblock \emph{arXiv preprint arXiv:2604.18095}.

\bibitem[{Miltiadous et~al.(2023)Miltiadous, Tzimourta, Afrantou, Ioannidis, Grigoriadis, Tsalikakis, Angelidis, Tsipouras, Glavas, Giannakeas et~al.}]{ADFTD}
Miltiadous, A.; Tzimourta, K.~D.; Afrantou, T.; Ioannidis, P.; Grigoriadis, N.; Tsalikakis, D.~G.; Angelidis, P.; Tsipouras, M.~G.; Glavas, E.; Giannakeas, N.; et~al. 2023.
\newblock {A dataset of scalp EEG recordings of Alzheimer’s disease, frontotemporal dementia and healthy subjects from routine EEG}.
\newblock \emph{Data}, 8(6): 95.

\bibitem[{Mumtaz et~al.(2017)Mumtaz, Xia, Ali, Yasin, Hussain, and Malik}]{Mumtaz2017}
Mumtaz, W.; Xia, L.; Ali, S. S.~A.; Yasin, M. A.~M.; Hussain, M.; and Malik, A.~S. 2017.
\newblock {Electroencephalogram (EEG)-based computer-aided technique to diagnose major depressive disorder (MDD)}.
\newblock \emph{Biomedical Signal Processing and Control}, 31: 108--115.

\bibitem[{Parsa et~al.(2023)Parsa, Rad, Vaezi, Hossein-Zadeh, Setarehdan, Rostami, Rostami, and Vahabie}]{parsa2023eeg}
Parsa, M.; Rad, H.~Y.; Vaezi, H.; Hossein-Zadeh, G.-A.; Setarehdan, S.~K.; Rostami, R.; Rostami, H.; and Vahabie, A.-H. 2023.
\newblock {EEG-based classification of individuals with neuropsychiatric disorders using deep neural networks: A systematic review of current status and future directions}.
\newblock \emph{Computer Methods and Programs in Biomedicine}, 240: 107683.

\bibitem[{Rockhill et~al.(2021)Rockhill, Jackson, George, Aron, and Swann}]{Rockhill2021}
Rockhill, A.~P.; Jackson, N.; George, J.; Aron, A.; and Swann, N.~C. 2021.
\newblock {UC San Diego Resting State EEG Data from Patients with Parkinson's Disease}.
\newblock OpenNeuro.

\bibitem[{Sadatnejad et~al.(2019)Sadatnejad, Rahmati, Rostami, Kazemi, Ghidary, M{\"u}ller, and Alimardani}]{sadatnejad2019eeg}
Sadatnejad, K.; Rahmati, M.; Rostami, R.; Kazemi, R.; Ghidary, S.~S.; M{\"u}ller, A.; and Alimardani, F. 2019.
\newblock {EEG representation using multi-instance framework on the manifold of symmetric positive definite matrices}.
\newblock \emph{Journal of Neural Engineering}, 16(3): 036016.

\bibitem[{Song et~al.(2022)Song, Zheng, Liu, and Gao}]{Conformer}
Song, Y.; Zheng, Q.; Liu, B.; and Gao, X. 2022.
\newblock {EEG conformer: Convolutional transformer for EEG decoding and visualization}.
\newblock \emph{IEEE Transactions on Neural Systems and Rehabilitation Engineering}, 31: 710--719.

\bibitem[{Wang et~al.(2025)Wang, Zhao, Luo, Zhou, Jiang, Li, Li, and Pan}]{CBraMod}
Wang, J.; Zhao, S.; Luo, Z.; Zhou, Y.; Jiang, H.; Li, S.; Li, T.; and Pan, G. 2025.
\newblock {Cbramod: A criss-cross brain foundation model for eeg decoding}.
\newblock In \emph{International conference on learning representations}, volume 2025, 75310--75346.

\bibitem[{Wang et~al.(2018)Wang, Yan, Tang, Bai, and Liu}]{Majority-Vote}
Wang, X.; Yan, Y.; Tang, P.; Bai, X.; and Liu, W. 2018.
\newblock {Revisiting multiple instance neural networks}.
\newblock \emph{Pattern recognition}, 74: 15--24.

\bibitem[{Wang et~al.(2023)Wang, Han, Wang, and Zhang}]{COMET}
Wang, Y.; Han, Y.; Wang, H.; and Zhang, X. 2023.
\newblock {Contrast everything: A hierarchical contrastive framework for medical time-series}.
\newblock \emph{Advances in Neural Information Processing Systems}, 36: 55694--55717.

\bibitem[{Wang et~al.(2024)Wang, Huang, Li, Yan, and Zhang}]{Medformer}
Wang, Y.; Huang, N.; Li, T.; Yan, Y.; and Zhang, X. 2024.
\newblock {Medformer: A multi-granularity patching transformer for medical time-series classification}.
\newblock \emph{Advances in Neural Information Processing Systems}, 37: 36314--36341.

\bibitem[{Weng et~al.(2026)Weng, Gu, Ma, Liu, Zhang, and Chen}]{State-Mamba}
Weng, W.; Gu, Y.; Ma, Y.; Liu, Y.; Zhang, Y.; and Chen, Y. 2026.
\newblock {State Mamba: Spatiotemporal EEG State-Space Model with Dynamic Brain Alignment for Cross-Subject Representation}.
\newblock In \emph{Proceedings of the AAAI Conference on Artificial Intelligence}, volume~40, 17850--17858.

\bibitem[{Xiao et~al.(2025)Xiao, Qi, Wang, He, Yu, Wu, Yu, Li, Gu, and Yu}]{EmotionMIL}
Xiao, J.; Qi, F.; Wang, L.; He, Y.; Yu, J.; Wu, W.; Yu, Z.; Li, Y.; Gu, Z.; and Yu, T. 2025.
\newblock {EmotionMIL: An End-to-End Multiple Instance Learning Framework for Emotion Recognition From EEG Signals}.
\newblock \emph{IEEE Transactions on Affective Computing}.

\bibitem[{Ye et~al.(2026)Ye, Zhang, Li, Li, and Tsung}]{MedSpaformer}
Ye, J.; Zhang, W.; Li, Z.; Li, J.; and Tsung, F. 2026.
\newblock {MedSpaformer: a Transferable Transformer with Multi-granularity Token Sparsification for Medical Time Series Classification}.
\newblock In \emph{Proceedings of the AAAI Conference on Artificial Intelligence}, volume~40, 27791--27799.

\bibitem[{Yu et~al.(2026)Yu, Wang, Yang, Qin, Aviles-Rivero, and Wang}]{CoTAR}
Yu, G.; Wang, J.; Yang, C.; Qin, J.; Aviles-Rivero, A.~I.; and Wang, S. 2026.
\newblock {Decentralized attention fails centralized signals: Rethinking transformers for medical time series}.
\newblock \emph{arXiv preprint arXiv:2602.18473}.

\bibitem[{Zerveas et~al.(2021)Zerveas, Jayaraman, Patel, Bhamidipaty, and Eickhoff}]{Masked-reconstruction}
Zerveas, G.; Jayaraman, S.; Patel, D.; Bhamidipaty, A.; and Eickhoff, C. 2021.
\newblock {A transformer-based framework for multivariate time series representation learning}.
\newblock In \emph{Proceedings of the 27th ACM SIGKDD conference on knowledge discovery \& data mining}, 2114--2124.

\bibitem[{Zini et~al.(2026)Zini, Barbera, Bianco, and Napoletano}]{MTDNet}
Zini, S.; Barbera, T.; Bianco, S.; and Napoletano, P. 2026.
\newblock {Alzheimer’s disease classification from EEG using a multiscale temporal deep network}.
\newblock \emph{Biomedical Signal Processing and Control}, 114: 109321.

\end{thebibliography}


\begin{thebibliography}{15}
\providecommand{\natexlab}[1]{#1}

\bibitem[{Chen et~al.(2024)Chen, Qiu, Zhu, Li, Wang, Sotiras, Wang, and Razi}]{TimeMIL}
Chen, X.; Qiu, P.; Zhu, W.; Li, H.; Wang, H.; Sotiras, A.; Wang, Y.; and Razi, A. 2024.
\newblock {TimeMIL: advancing multivariate time series classification via a time-aware multiple instance learning}.
\newblock In \emph{Proceedings of the 41st International Conference on Machine Learning}, 7190--7206.

\bibitem[{Early et~al.(2024)Early, Cheung, Cutajar, Xie, Kandola, and Twomey}]{MILLET}
Early, J.; Cheung, G.; Cutajar, K.; Xie, H.; Kandola, J.; and Twomey, N. 2024.
\newblock {Inherently interpretable time series classification via multiple instance learning}.
\newblock In \emph{International Conference on Learning Representations}, volume 2024, 56800--56828.

\bibitem[{Ilse, Tomczak, and Welling(2018)}]{Attention}
Ilse, M.; Tomczak, J.; and Welling, M. 2018.
\newblock {Attention-based deep multiple instance learning}.
\newblock In \emph{International conference on machine learning}, 2127--2136. PMLR.

\bibitem[{Javed et~al.(2022)Javed, Juyal, Padigela, Taylor-Weiner, Yu, and Prakash}]{Additive}
Javed, S.~A.; Juyal, D.; Padigela, H.; Taylor-Weiner, A.; Yu, L.; and Prakash, A. 2022.
\newblock {Additive mil: Intrinsically interpretable multiple instance learning for pathology}.
\newblock \emph{Advances in Neural Information Processing Systems}, 35: 20689--20702.

\bibitem[{Kachare et~al.(2024)Kachare, Puri, Sangle, Al-Shourbaji, Jabbari, Kirner, Alameen, Migdady, and Abualigah}]{LCADNet}
Kachare, P.; Puri, D.; Sangle, S.~B.; Al-Shourbaji, I.; Jabbari, A.; Kirner, R.; Alameen, A.; Migdady, H.; and Abualigah, L. 2024.
\newblock {LCADNet: a novel light CNN architecture for EEG-based Alzheimer disease detection}.
\newblock \emph{Physical and Engineering Sciences in Medicine}, 47(3): 1037--1050.

\bibitem[{Khosla et~al.(2020)Khosla, Teterwak, Wang, Sarna, Tian, Isola, Maschinot, Liu, and Krishnan}]{SupCon}
Khosla, P.; Teterwak, P.; Wang, C.; Sarna, A.; Tian, Y.; Isola, P.; Maschinot, A.; Liu, C.; and Krishnan, D. 2020.
\newblock {Supervised contrastive learning}.
\newblock \emph{Advances in neural information processing systems}, 33: 18661--18673.

\bibitem[{Lawhern et~al.(2018)Lawhern, Solon, Waytowich, Gordon, Hung, and Lance}]{EEGNet}
Lawhern, V.~J.; Solon, A.~J.; Waytowich, N.~R.; Gordon, S.~M.; Hung, C.~P.; and Lance, B.~J. 2018.
\newblock {EEGNet: a compact convolutional neural network for EEG-based brain--computer interfaces}.
\newblock \emph{Journal of neural engineering}, 15(5): 056013.

\bibitem[{Ma et~al.(2026)Ma, Li, Qiu, Li, Meng, Zhang, Liu, Shen, and Song}]{DSAINet}
Ma, Z.; Li, Z.; Qiu, Z.; Li, J.; Meng, L.; Zhang, X.; Liu, Y.; Shen, X.; and Song, S. 2026.
\newblock {DSAINet: An Efficient Dual-Scale Attentive Interaction Network for General EEG Decoding}.
\newblock \emph{arXiv preprint arXiv:2604.18095}.

\bibitem[{Miltiadous et~al.(2023)Miltiadous, Tzimourta, Afrantou, Ioannidis, Grigoriadis, Tsalikakis, Angelidis, Tsipouras, Glavas, Giannakeas et~al.}]{ADFTD}
Miltiadous, A.; Tzimourta, K.~D.; Afrantou, T.; Ioannidis, P.; Grigoriadis, N.; Tsalikakis, D.~G.; Angelidis, P.; Tsipouras, M.~G.; Glavas, E.; Giannakeas, N.; et~al. 2023.
\newblock {A dataset of scalp EEG recordings of Alzheimer’s disease, frontotemporal dementia and healthy subjects from routine EEG}.
\newblock \emph{Data}, 8(6): 95.

\bibitem[{Mumtaz et~al.(2017)Mumtaz, Xia, Ali, Yasin, Hussain, and Malik}]{Mumtaz2017}
Mumtaz, W.; Xia, L.; Ali, S. S.~A.; Yasin, M. A.~M.; Hussain, M.; and Malik, A.~S. 2017.
\newblock {Electroencephalogram (EEG)-based computer-aided technique to diagnose major depressive disorder (MDD)}.
\newblock \emph{Biomedical Signal Processing and Control}, 31: 108--115.

\bibitem[{Rockhill et~al.(2021)Rockhill, Jackson, George, Aron, and Swann}]{Rockhill2021}
Rockhill, A.~P.; Jackson, N.; George, J.; Aron, A.; and Swann, N.~C. 2021.
\newblock {UC San Diego Resting State EEG Data from Patients with Parkinson's Disease}.
\newblock OpenNeuro.

\bibitem[{Song et~al.(2022)Song, Zheng, Liu, and Gao}]{Conformer}
Song, Y.; Zheng, Q.; Liu, B.; and Gao, X. 2022.
\newblock {EEG conformer: Convolutional transformer for EEG decoding and visualization}.
\newblock \emph{IEEE Transactions on Neural Systems and Rehabilitation Engineering}, 31: 710--719.

\bibitem[{Wang et~al.(2018)Wang, Yan, Tang, Bai, and Liu}]{Majority-Vote}
Wang, X.; Yan, Y.; Tang, P.; Bai, X.; and Liu, W. 2018.
\newblock {Revisiting multiple instance neural networks}.
\newblock \emph{Pattern recognition}, 74: 15--24.

\bibitem[{Zerveas et~al.(2021)Zerveas, Jayaraman, Patel, Bhamidipaty, and Eickhoff}]{Masked-reconstruction}
Zerveas, G.; Jayaraman, S.; Patel, D.; Bhamidipaty, A.; and Eickhoff, C. 2021.
\newblock {A transformer-based framework for multivariate time series representation learning}.
\newblock In \emph{Proceedings of the 27th ACM SIGKDD conference on knowledge discovery \& data mining}, 2114--2124.

\bibitem[{Zini et~al.(2026)Zini, Barbera, Bianco, and Napoletano}]{MTDNet}
Zini, S.; Barbera, T.; Bianco, S.; and Napoletano, P. 2026.
\newblock {Alzheimer’s disease classification from EEG using a multiscale temporal deep network}.
\newblock \emph{Biomedical Signal Processing and Control}, 114: 109321.

\end{thebibliography}

\newpage
\section{A. Datasets}

We adopted three public EEG disease datasets. 

\begin{itemize}
    \item ADFTD \citesupp{ADFTD}: This resting-state EEG dataset contains 36 subjects with Alzheimer’s disease (AD), 23 subjects with frontotemporal dementia (FTD), and 29 age-matched healthy controls (HC). EEG was recorded from 19 scalp channels while participants rested with their eyes closed. After preprocessing and non-overlapping 1-s segmentation, the dataset yields 69,794 instances.
    \item Mumtaz2017 \citesupp{Mumtaz2017}: This dataset comprises 34 subjects with major depressive disorder (MDD) and 30 healthy controls (HC), with 19-channel EEG recorded during eyes-open (EO), eyes-closed (EC), and task sessions. We retain only the EO and EC recordings and exclude one MDD subject because of its lack of these recordings, leaving 63 subjects and 35,891 non-overlapping 1-s instances.
    \item Rockhill2021 \citesupp{Rockhill2021}: This dataset contains resting-state EEG recordings from 15 subjects with Parkinson’s disease (PD) and 16 healthy controls (HC), acquired using 32 EEG channels. For participants with PD, we use only the medication-ON recordings. With 50\% overlapping 1-s segmentation, the dataset yields 12,033 instances.
\end{itemize}

\section{B. Backbones}

To evaluate the framework across different EEG decoding architectures, we use five representative backbones.

\begin{itemize}
    \item EEGNet \citesupp{EEGNet}: EEGNet is a compact CNN designed for efficient EEG decoding. It first learns temporal frequency-selective filters, then applies depthwise spatial convolution across EEG channels, followed by separable convolution to integrate the resulting spatiotemporal features with a small number of parameters.
    \item Conformer \citesupp{Conformer}: EEG Conformer is a compact CNN–Transformer architecture that combines local and global EEG modeling. Its convolutional module extracts low-level temporal and spatial patterns, while Transformer self-attention captures long-range dependencies among the resulting EEG tokens.
    \item LCADNet \citesupp{LCADNet}: LCADNet is a lightweight CNN originally developed for EEG-based Alzheimer’s disease detection. It uses two convolutional layers to extract disease-related EEG patterns, with intermediate max pooling to reduce temporal redundancy, followed by compact fully connected layers for discrimination.
    \item DSAINet \citesupp{DSAINet}: DSAINet is an efficient dual-scale architecture for general EEG decoding. It models fine- and coarse-scale temporal dynamics through parallel convolutional branches, refines scale-specific patterns with intra-branch attention, and integrates complementary information through inter-branch attention and adaptive token aggregation.
    \item MTDNet \citesupp{MTDNet}: MTDNet is a lightweight multiscale temporal network developed for EEG-based Alzheimer’s disease classification. It combines temporal convolutions with different receptive fields and recurrent modeling to jointly capture short-term signal patterns and longer-range EEG dynamics.
\end{itemize}

\section{C. Baselines}
We compare our framework with four groups of baselines. 

\textbf{1. Majority Vote}
    \begin{itemize}
        \item An instance classifier is trained by assigning the corresponding subject diagnosis to every EEG instance. At inference, predictions from all instances belonging to the same subject are aggregated by majority voting to obtain the final subject-level diagnosis. \citesupp{Majority-Vote}
    \end{itemize}
     
\textbf{2. GenericMIL methods}
\begin{itemize}
    \item Attention-based MIL \citesupp{Attention}: Attention-based MIL learns a permutation-invariant attention function that assigns a relevance weight to each instance embedding. The weighted instance representations are aggregated into a bag representation, which is then used to predict the subject label while requiring only bag-level supervision.
    \item Additive MIL \citesupp{Additive}: Additive MIL reformulates bag prediction as the sum of class-specific contributions from individual instances. Each attended instance is mapped to a prediction contribution before aggregation, allowing the final bag-level output to be exactly decomposed into positive or negative instance-level evidence.
\end{itemize}

\textbf{3. ModernMIL methods}
\begin{itemize}
    \item MILLET \citesupp{MILLET}: MILLET reformulates time-series classification as an MIL problem by treating temporal positions or local features as instances. It combines instance-level prediction with MIL pooling to identify localized evidence within a time series and provide inherently sparse, instance-level explanations for the final classification.
    \item TimeMIL \citesupp{TimeMIL}: TimeMIL is a time-aware MIL framework designed to preserve temporal ordering and dependencies among instances. It uses a tokenized Transformer together with learnable wavelet positional encoding to locate informative temporal patterns and aggregate them through time-aware MIL pooling.
\end{itemize}

\textbf{4. Two Stage Representation Learning}
\begin{itemize}
    \item SupCon \citesupp{SupCon}: Supervised contrastive learning pulls representations carrying the same class label closer together while separating representations from different classes. In our implementation, subject diagnoses are inherited by Stage-1 EEG instances to construct positive and negative sets, after which the pretrained encoder is transferred to the same attention-based MIL stage.
    \item Masked Reconstruction \citesupp{Masked-reconstruction}: Masked reconstruction randomly hides portions of each EEG instance and trains the encoder to recover the missing signal values from the remaining temporal and cross-channel context. The resulting pretrained encoder is subsequently transferred to the same attention-based MIL stage used by our framework.
\end{itemize}

\section{D. Computing Infrastructure and Implementation Details}

\paragraph{Computing infrastructure.}
Experiments were run on a Linux workstation with eight NVIDIA GeForce RTX 4090 GPUs, each with 24 GB of GPU memory, and two AMD EPYC 7542 32-core processors, providing 64 physical CPU cores and 128 hardware threads. The system had 251 GiB of RAM and ran Ubuntu 20.04.6 LTS with Linux kernel 5.15.0. The software environment used Python 3.10.20 and PyTorch 2.5.1 built with CUDA 12.1. Additional dependencies included einops, LMDB, MNE, MNE-BIDS, NumPy, PyYAML, scikit-learn, and SciPy.

\paragraph{Method implementations.}
All supervised classification objectives use cross-entropy loss. Variable-length subject bags are zero-padded to the longest bag within each mini-batch and accompanied by a Boolean validity mask. Padded instances are excluded from both backbone encoding and MIL aggregation. Channel-wise z-score statistics are estimated exclusively from the training split and then applied unchanged to the corresponding validation and test splits.

For the \emph{majority-voting} baseline, the backbone and its instance-level classifier are trained using the diagnosis inherited from the corresponding subject. Training uses Adam for 100 epochs with a batch size of 512, a learning rate of $5\times10^{-4}$, and weight decay of $10^{-4}$. At inference, the predicted classes of all instances from a subject are aggregated by majority vote. When multiple classes receive the same number of votes, the class with the highest mean predicted probability is selected.

The end-to-end MIL baselines are trained for 100 epochs using Adam with eight subjects per mini-batch, a learning rate of $5\times10^{-4}$, and weight decay of $10^{-4}$. Each instance is first encoded independently by the selected EEG backbone. For attention-based MIL~\citesupp{Attention}, sinusoidal positional encoding and dropout with probability $0.1$ are applied to the instance embeddings. An attention network with an eight-dimensional hidden layer, $\tanh$ activation, and sigmoid output produces one gate per instance. The gated instance embeddings are mean-aggregated and passed to a linear bag classifier. Additive MIL~\citesupp{Additive} instead maps the gated embeddings to instance-level logits and averages these logits to obtain the bag prediction. Our MILLET implementation~\citesupp{MILLET} follows the conjunctive formulation: each ungated embedding is first mapped to class logits, after which the logits are modulated by the learned instance gates and averaged across the bag.

The TimeMIL baseline~\citesupp{TimeMIL} preserves the temporal ordering of the windows. Instance embeddings are projected to a model dimension of $\min(d,512)$, where $d$ is the backbone output dimension, and are combined with a learnable classification token. The aggregator contains two transformer blocks with dropout $0.1$ and a feed-forward expansion ratio of four. Before each transformer block, the instance sequence is augmented by a learnable wavelet positional encoding comprising three depth-wise Mexican-hat wavelets with kernel size 19. The number of attention heads is chosen as the largest divisor of the model dimension not exceeding eight. The final classification-token representation is processed by a two-layer MLP to produce the subject prediction.

BridgeMIL, SupCon, and masked reconstruction share the same two-stage optimization protocol. Stage~1 is trained for 50 epochs using AdamW with a batch size of 512, an initial learning rate of $10^{-3}$, weight decay of $10^{-4}$, five linear warm-up epochs, and cosine learning-rate decay. After pretraining, the Stage~1 projection or reconstruction head is discarded, and only the pretrained backbone encoder is transferred to Stage~2.

For \emph{BridgeMIL}, two types of positive pairs are sampled in every Stage~1 iteration. First, a window is paired with another window from the same subject located at most two temporal steps away. Second, two sub-bags are independently sampled from the same subject, each containing eight windows. A sub-bag representation is obtained by averaging the projected representations of its constituent windows. Both pair types use the same projection head:
\begin{equation}
d \rightarrow 256 \rightarrow 128,
\end{equation}
consisting of a linear layer, layer normalization, ELU activation, and a second linear layer. For each pair $(\mathbf{z},\mathbf{z}')$, the objective is
\begin{equation}
\mathcal{L}_{\mathrm{pair}}
=
25\mathcal{L}_{\mathrm{inv}}
+
25\mathcal{L}_{\mathrm{var}}
+
\mathcal{L}_{\mathrm{cov}},
\label{eq:bridge_pair_loss}
\end{equation}
where $\mathcal{L}_{\mathrm{inv}}$ is the mean-squared difference between the paired representations. The variance term penalizes feature dimensions whose batch standard deviation is below one, using $\epsilon=10^{-4}$ for numerical stability. The covariance term penalizes the squared off-diagonal entries of the covariance matrices of both views. The nearby-window and sub-bag losses are averaged to form the final Stage~1 objective.

For the \emph{SupCon} baseline~\citesupp{SupCon}, all instances retain their inherited subject diagnoses during Stage~1. Mini-batches are sampled using inverse-frequency class weights to reduce class imbalance. Two independently augmented views are produced for each instance using Gaussian jitter, amplitude scaling, and circular temporal shifting. Each augmentation is applied with probability $0.5$. The jitter standard deviation is $0.05$ times the temporal standard deviation of each channel, the scaling factor is sampled uniformly from $[0.8,1.2]$, and the maximum temporal shift is $5\%$ of the window length. The same $256$--$128$ projection head used by BridgeMIL is employed, and the supervised contrastive temperature is set to $0.07$. Instances carrying the same inherited diagnosis are treated as positives.

For the \emph{masked-reconstruction} baseline~\citesupp{Masked-reconstruction}, approximately $15\%$ of the temporal samples in each channel are replaced by zero. Masked positions are generated as contiguous spans of 16 samples, corresponding to 80 ms at the sampling rate of 200 Hz. The encoder output is passed to an MLP reconstruction head that predicts the complete multichannel window. Its hidden dimension is
\begin{equation}
d_{\mathrm{hidden}}
=
\min(512,4d),
\label{eq:reconstruction_hidden_dimension}
\end{equation}
where $d$ is the backbone representation dimension. Reconstruction is optimized using mean-squared error evaluated only at the masked positions.

Stage~2 uses the same attention-based MIL head for all three two-stage methods. It is trained for 100 epochs using AdamW with eight subjects per mini-batch, five warm-up epochs, and cosine learning-rate decay. The encoder and MIL head use learning rates of $10^{-4}$ and $5\times10^{-4}$, respectively, with weight decay $10^{-4}$. The encoder is frozen during the first five epochs. A frozen copy of the Stage~1 encoder serves as the feature-retention reference. For each Stage~2 mini-batch, the retention loss is the mean-squared difference between the current and reference instance representations:
\begin{equation}
\mathcal{L}_{\mathrm{S2}}
=
\mathcal{L}_{\mathrm{CE}}
+
\lambda_{\mathrm{ret}}\mathcal{L}_{\mathrm{ret}},
\qquad
\lambda_{\mathrm{ret}}=10^{-5}.
\label{eq:stage2_objective}
\end{equation}
Retention is applied only at the instance level; no bag-level representation is constrained. When a mini-batch contains more than 512 valid instances, 512 are sampled uniformly for computing $\mathcal{L}_{\mathrm{ret}}$.

For all supervised and Stage~2 experiments, the checkpoint with the highest validation accuracy is retained, with lower validation loss used to break ties. To avoid selecting transient checkpoints from the beginning of training, validation-based checkpoint selection starts after epoch 10 for ADFTD and after epoch 15 for Rockhill2021; no additional burn-in is used for Mumtaz2017.

\paragraph{Backbone implementations.}
All backbones receive windows in the format $\mathbb{R}^{1\times C\times T}$, where $C$ is the dataset-specific number of EEG channels and $T=200$ for a one-second window. For majority voting, the complete backbone-specific classifier is used. For MIL and Stage~1 pretraining, the original classification layer is removed and the representation immediately preceding it is exposed as the instance embedding.

\noindent\textbf{EEGNet~\citesupp{EEGNet}.}
EEGNet uses an initial temporal convolution with kernel length 64 and $F_1=8$ filters, followed by a depth-wise spatial convolution across all EEG channels with depth multiplier $D=2$. The first block uses average pooling by a factor of four. The second block contains a depth-wise separable temporal convolution with kernel length 16, $F_2=16$ output filters, and average pooling by a factor of eight. ELU activations, batch normalization, and dropout with probability $0.25$ are used throughout. The flattened output of the second convolutional block is used as the instance representation.

\noindent\textbf{Conformer~\citesupp{Conformer}.}
The Conformer begins with a shallow convolutional patch embedding. It applies a temporal convolution with kernel size 25, a spatial convolution spanning all channels, and average pooling with temporal kernel size 75 and stride 15. The resulting tokens have dimension 40. Four transformer encoder blocks are used, each with ten attention heads and a feed-forward expansion ratio of four. Dropout is set to $0.25$ in the patch embedding, self-attention, and feed-forward layers. For MIL and pretraining, all transformer tokens are flattened to form the instance representation. The standalone instance classifier additionally uses an MLP with hidden dimensions 256 and 32.

\noindent\textbf{LCADNet~\citesupp{LCADNet}.}
LCADNet contains two two-dimensional convolutional layers with 20 and 10 output channels, respectively. Both use $3\times3$ kernels and ReLU activation, with a $2\times21$ max-pooling layer after the first convolution. The standalone classifier uses fully connected layers of sizes 50 and 80, with dropout $0.25$. For MIL and pretraining, the convolutional feature map is flattened before the fully connected classifier and used as the instance representation.

\noindent\textbf{DSAINet~\citesupp{DSAINet}.}
DSAINet first uses an EEGNet-style patch embedding with $F_1=16$, depth multiplier $D=2$, temporal kernel size 64, pooling factors of four and eight, and dropout $0.25$. The resulting tokens are projected to dimension 40 and combined with learnable positional embeddings. Two temporal convolution branches use kernel sequences $\{11,15\}$ and $\{3,7\}$, respectively, with expansion factor four. Each branch is processed by one four-head self-attention block, followed by bidirectional cross-attention between the two branches. The self- and cross-attention feed-forward expansion ratios are both two, and attention dropout is $0.25$. Learnable residual connections from the shared token representation to both branches are enabled. Attention pooling is performed independently within each branch, and the two pooled 40-dimensional representations are concatenated into an 80-dimensional instance embedding.

\noindent\textbf{MTDNet~\citesupp{MTDNet}.}
MTDNet processes each window through three parallel temporal branches. The branches use convolution kernel and stride pairs $(10,10)$, $(5,5)$, and $(2,2)$, respectively, with 32 output channels per branch. The latter two branches additionally use average-pooling factors of two and five so that the three temporal resolutions can be concatenated. The concatenated sequence is processed by a two-layer unidirectional LSTM with hidden dimension 16. For MIL and pretraining, the batch-normalized final LSTM state is used as the 16-dimensional instance representation. The standalone classifier adds a 16-dimensional fully connected layer before the output layer.

\section{E. Ablation}
\label{sec:supp_ablation}

\begin{table*}[t]
\centering
{\small
\setlength{\tabcolsep}{3.2pt}
\renewcommand{\arraystretch}{1.08}
\begin{tabular}{lcccccc}
\toprule
\multirow{3}{*}{Dataset}
& \multicolumn{4}{c}{Stage~1 Alternatives and Ablations}
& \multicolumn{1}{c}{Stage~2 Ablation}
& \multicolumn{1}{c}{Ours} \\
\cmidrule(lr){2-5}
\cmidrule(lr){6-6}
\cmidrule(lr){7-7}
&
\multirow{2}{*}{\makecell{Supervised\\Stage~1}}
&
\multirow{2}{*}{\makecell{w/o Variance\\\& Covariance}}
&
\multirow{2}{*}{\makecell{w/o Sub-bag}}
&
\multirow{2}{*}{\makecell{w/o Nearby}}
&
\multirow{2}{*}{\makecell{w/o Feature\\Retention}}
&
\multirow{2}{*}{\makecell{Full\\BridgeMIL}} \\
& & & & & & \\
\midrule

ADFTD
& 60.80 $\pm$ 0.28
& 45.12 $\pm$ 1.45
& 60.87 $\pm$ 1.66
& 58.49 $\pm$ 1.62
& 60.37 $\pm$ 0.85
& \textbf{63.40 $\pm$ 0.95} \\

\midrule

Mumtaz2017
& 89.14 $\pm$ 0.47
& 71.48 $\pm$ 0.99
& 86.71 $\pm$ 1.52
& 86.86 $\pm$ 1.20
& 85.00 $\pm$ 2.34
& \textbf{91.51 $\pm$ 1.46} \\

\midrule

Rockhill2021
& 74.19 $\pm$ 2.19
& 71.81 $\pm$ 3.16
& 75.00 $\pm$ 2.31
& 72.33 $\pm$ 1.84
& 72.29 $\pm$ 1.11
& \textbf{77.38 $\pm$ 2.10} \\

\bottomrule
\end{tabular}
}
\caption{
Ablation results using EEGNet on three EEG disease datasets. Results
report subject-level classification accuracy (mean $\pm$ standard
deviation across five seeds, \%). 
The complete BridgeMIL results are shown in bold.
}
\label{tab:supp_ablation}
\end{table*}

We conduct an ablation study with EEGNet on all three datasets to examine the contributions of the Stage~1 objective and Stage~2 feature retention. All variants follow the same data splits and training protocol, and results are reported as subject-level accuracy across five seeds. \emph{Supervised Stage~1} uses inherited instance labels for pretraining; \emph{w/o Variance \& Covariance} removes the corresponding regularization terms; \emph{w/o Sub-bag} and \emph{w/o Nearby} retain only one of the two alignment scales. The Stage~2 variant removes feature retention while using the same pretrained encoder as BridgeMIL.

As shown in Table~\ref{tab:supp_ablation}, the complete BridgeMIL framework achieves the highest mean accuracy on all three datasets. Its consistent advantage over Supervised Stage~1 shows that the improvement is not explained solely by two-stage training; learning without inherited instance labels provides a better initialization for subject-level MIL.

Removing variance and covariance regularization causes the largest Stage~1 degradation, demonstrating their importance in preventing collapsed or redundant representations. Using either nearby-window or sub-bag alignment alone also consistently underperforms their combination. This supports their complementary roles: nearby pairs capture local temporal consistency, whereas sub-bags encode broader within-subject structure.

Finally, removing feature retention reduces accuracy by 3.03, 6.51, and 5.09 percentage points on ADFTD, Mumtaz2017, and Rockhill2021, respectively. Because this variant uses the same Stage~1 encoder, the reduction indicates that unrestricted Stage~2 fine-tuning can overwrite useful pretrained instance representations. Feature retention therefore provides an effective bridge between instance representation learning and subject-level supervision.


\section{F. Hyperparameter Sensitivity}
\label{sec:hyperparameter_sensitivity}

We examine the sensitivity of BridgeMIL to two hyperparameters: the
feature-retention coefficient $\lambda_{\mathrm{ret}}$ in Stage~2 and the
sub-bag size $K$ used in Stage~1. This analysis
is conducted on ADFTD with EEGNet. We vary one hyperparameter at a time while
holding the other at the setting used in the main experiments, namely
$\lambda_{\mathrm{ret}}=10^{-5}$ and $K=8$. All results report subject-level
accuracy averaged over five random seeds.

\begin{table}[t]
    \centering
    \setlength{\tabcolsep}{8pt}
    \renewcommand{\arraystretch}{1.12}
    \begin{tabular}{lcc}
        \toprule
        Hyperparameter & Value & Accuracy (\%) \\
        \midrule
        \multicolumn{3}{l}{\textit{Feature-retention coefficient
        ($K=8$)}} \\
        $\lambda_{\mathrm{ret}}$ & $10^{-6}$
            & $61.30 \pm 0.82$ \\
        $\lambda_{\mathrm{ret}}$ & $\mathbf{10^{-5}}$
            & $\mathbf{63.40 \pm 0.95}$ \\
        $\lambda_{\mathrm{ret}}$ & $10^{-4}$
            & $62.63 \pm 1.30$ \\
        \midrule
        \multicolumn{3}{l}{\textit{Sub-bag size
        ($\lambda_{\mathrm{ret}}=10^{-5}$)}} \\
        $K$ & $4$
            & $62.13 \pm 1.90$ \\
        $K$ & $\mathbf{8}$
            & $\mathbf{63.40 \pm 0.95}$ \\
        $K$ & $16$
            & $61.63 \pm 1.22$ \\
        \bottomrule
    \end{tabular}
    \caption{Hyperparameter sensitivity of BridgeMIL with EEGNet on ADFTD.
    We vary one hyperparameter at a time while fixing the other to the setting
    used in the main experiments. Results are subject-level accuracy
    (\%, mean $\pm$ standard deviation) over five seeds. Bold parameter values
    denote the settings used in the main experiments, and bold results indicate
    the best performance within each group.}
    \label{tab:hyp}
\end{table}

As shown in Table~\ref{tab:hyp}, the configuration used
in the main experiments achieves the highest mean accuracy among the tested
values. For feature retention, decreasing $\lambda_{\mathrm{ret}}$ from
$10^{-5}$ to $10^{-6}$ reduces the accuracy from $63.40\%$ to $61.30\%$,
whereas increasing it to $10^{-4}$ yields $62.63\%$. This suggests that an
intermediate retention strength better balances preservation of the pretrained
instance representations with adaptation to subject-level supervision.

Similarly, sub-bag sizes of $4$ and $16$ achieve accuracies of $62.13\%$ and
$61.63\%$, respectively, compared with $63.40\%$ for $K=8$. Overall, the
results remain reasonably consistent across the tested values, while the
hyperparameters adopted in the main experiments provide the best performance.

\bibliographystylesupp{aaai2027}
\bibliographysupp{supp}

\end{document}